%% file: main.tex
\documentclass{article}
\usepackage[nonatbib,preprint]{neurips_2020}

\usepackage[utf8]{inputenc} % allow utf-8 input
\usepackage[T1]{fontenc}    % use 8-bit T1 fonts
\usepackage{hyperref}       % hyperlinks
\usepackage{url}            % simple URL typesetting
\usepackage{booktabs}       % professional-quality tables
\usepackage{amsfonts}       % blackboard math symbols
\usepackage{nicefrac}       % compact symbols for 1/2, etc.
\usepackage{microtype}      % microtypography

\input{tex/plots}

\usepackage{epsfig}
\usepackage{graphicx}
\usepackage{amsmath}
\usepackage{amssymb}
\usepackage{xspace}
\usepackage{bbm}
\usepackage{multirow}
\usepackage{enumitem}
\usepackage{lipsum}

\hypersetup{colorlinks=true}
\hypersetup{pageanchor=false}
\usepackage[ruled,vlined,linesnumbered]{algorithm2e}

\title{Asymmetric metric learning for knowledge transfer}

\author{
Mateusz Budnik \quad Yannis Avrithis \\
{\fontsize{11}{13}\selectfont Inria, IRISA, Univ Rennes, CNRS} \\
}

\begin{document}

\maketitle

\input{tex/abbrev}
\input{tex/defn}

\input{tex/defn_exp}

\begin{abstract}

Knowledge transfer from large teacher models to smaller student models has recently been studied for metric learning, focusing on fine-grained classification. In this work, focusing on instance-level image retrieval, we study an \emph{asymmetric testing} task, where the database is represented by the teacher and queries by the student. Inspired by this task, we introduce \emph{asymmetric metric learning}, a novel paradigm of using asymmetric representations at \emph{training}. This acts as a simple combination of knowledge transfer with the original metric learning task.

We systematically evaluate different teacher and student models, metric learning and knowledge transfer loss functions
on the new asymmetric testing as well as the standard \emph{symmetric testing} task, where database and queries are represented by the same model. We find that plain \emph{regression} is surprisingly effective compared to more complex knowledge transfer mechanisms, working best in asymmetric testing. Interestingly, our asymmetric metric learning approach works best in symmetric testing, allowing the student to even outperform the teacher.

\end{abstract}

\input{tex/intro}

\input{tex/related}

\input{tex/method}

\input{tex/exp_setup}

\input{tex/exp_results}
\input{tex/conclusion}

\clearpage

{\small
  \bibliographystyle{ieee_fullname}
  \bibliography{main}
}

 \clearpage
 \input{tex/appendix}

\end{document}

%% file: tex/plots.tex
\usepackage{tikz}
\usetikzlibrary{arrows.meta,shapes,calc,matrix,fit,positioning,backgrounds,decorations.markings}
\usepackage{pgfplots}
\usepackage{pgfplotstable}
\pgfplotsset{compat=1.9}

\usepgfplotslibrary{external}
\newcommand{\extfig}[2]{\tikzsetnextfilename{#1}{#2}}

\tikzstyle{every picture}+=[
	remember picture,
	every text node part/.style={align=center},
	every matrix/.append style={ampersand replacement=\&},
]
\tikzstyle{tight} = [inner sep=0pt,outer sep=0pt]
\tikzstyle{node}  = [draw,circle,tight,minimum size=12pt,anchor=center]
\tikzstyle{op}    = [draw,circle,tight]
\tikzstyle{dot}   = [fill,draw,circle,inner sep=1pt,outer sep=0]
\tikzstyle{pt}    = [fill,draw,circle,inner sep=1.5pt,outer sep=.2pt]
\tikzstyle{box}   = [draw,thick,rectangle,inner sep=3pt]
\tikzstyle{high}  = [black!60]
\tikzstyle{group} = [high,box,opacity=.5]
\tikzstyle{rectc} = [tight,transform shape]
\tikzstyle{rect}  = [rectc,anchor=south west]

\tikzset{every mark/.append style={solid}}
\pgfplotsset{%smooth,
	grid=both, width=\columnwidth, try min ticks=5,
	every axis/.append style={font=\small},
	every axis plot/.append style={thick,mark=none,mark size=1.8,tension=0.18},
	legend cell align=left, legend style={fill opacity=0.8},
	xticklabel={\pgfmathprintnumber[assume math mode=true]{\tick}},
	yticklabel={\pgfmathprintnumber[assume math mode=true]{\tick}},
	nodes near coords math/.style={
		nodes near coords={\pgfmathprintnumber[assume math mode=true]{\pgfplotspointmeta}},
	},
}

\pgfplotsset{
	dash/.style={mark=o,dashed,opacity=0.6},
	dott/.style={mark=o,dotted,opacity=0.6},
	nolim/.style={enlargelimits=false},
	plain/.style={every axis plot/.append style={},nolim,grid=none},
}

%% file: tex/abbrev.tex
\newcommand{\head}[1]{{\smallskip\noindent\textbf{#1}}}
\newcommand{\alert}[1]{{\color{black}{#1}}}
\newcommand{\eq}[1]{(\ref{eq:#1})}

\newcommand{\Th}[1]{\textsc{#1}}
\newcommand{\mr}[2]{\multirow{#1}{*}{#2}}
\newcommand{\mc}[2]{\multicolumn{#1}{c}{#2}}
\newcommand{\tb}[1]{\textbf{#1}}

\newcommand{\red}[1]{{\color{red}{#1}}}
\newcommand{\blue}[1]{{\color{blue}{#1}}}
\newcommand{\green}[1]{{\color{green}{#1}}}
\newcommand{\gray}[1]{{\color{gray}{#1}}}

\newcommand{\citeme}[1]{\red{[XX]}}
\newcommand{\refme}[1]{\red{(XX)}}

\newcommand{\fig}[2][1]{\includegraphics[width=#1\columnwidth]{fig/#2}}
\newcommand{\figh}[2][1]{\includegraphics[height=#1\columnwidth]{fig/#2}}
\newcommand{\figr}[2][1]{\includegraphics[width=#1\columnwidth,height=#1\columnwidth]{fig/#2}}

%--------------------------------------------------------------------

\newcommand{\tran}{^\top}
\newcommand{\mtran}{^{-\top}}
\newcommand{\zcol}{\mathbf{0}}
\newcommand{\zrow}{\zcol\tran}
\newcommand{\ch}{\checkmark}

\newcommand{\ind}{\mathbbm{1}}
\newcommand{\expect}{\mathbb{E}}
\newcommand{\nat}{\mathbb{N}}
\newcommand{\zahl}{\mathbb{Z}}
\newcommand{\real}{\mathbb{R}}
\newcommand{\proj}{\mathbb{P}}
\newcommand{\prob}{\mathbf{Pr}}
\newcommand{\normal}{\mathcal{N}}

\newcommand{\mif}{\textrm{if}\ }
\newcommand{\other}{\textrm{otherwise}}
\newcommand{\minimize}{\textrm{minimize}\ }
\newcommand{\maximize}{\textrm{maximize}\ }
\newcommand{\st}{\textrm{subject\ to}\ }

\newcommand{\id}{\operatorname{id}}
\newcommand{\const}{\operatorname{const}}
\newcommand{\sgn}{\operatorname{sgn}}
\newcommand{\var}{\operatorname{Var}}
\newcommand{\mean}{\operatorname{mean}}
\newcommand{\trace}{\operatorname{tr}}
\newcommand{\diag}{\operatorname{diag}}
\newcommand{\vect}{\operatorname{vec}}
\newcommand{\cov}{\operatorname{cov}}
\newcommand{\sign}{\operatorname{sign}}
\newcommand{\prj}{\operatorname{proj}}

\newcommand{\softmax}{\operatorname{softmax}}
\newcommand{\clip}{\operatorname{clip}}

\newcommand{\defn}{\mathrel{:=}}
\newcommand{\peq}{\mathrel{+\!=}}
\newcommand{\meq}{\mathrel{-\!=}}

\newcommand{\floor}[1]{\left\lfloor{#1}\right\rfloor}
\newcommand{\ceil}[1]{\left\lceil{#1}\right\rceil}
\newcommand{\inner}[1]{\left\langle{#1}\right\rangle}
\newcommand{\norm}[1]{\left\|{#1}\right\|}
\newcommand{\abs}[1]{\left|{#1}\right|}
\newcommand{\frob}[1]{\norm{#1}_F}
\newcommand{\card}[1]{\left|{#1}\right|\xspace}
\newcommand{\diff}{\mathrm{d}}
\newcommand{\der}[3][]{\frac{d^{#1}#2}{d#3^{#1}}}
\newcommand{\pder}[3][]{\frac{\partial^{#1}{#2}}{\partial{#3^{#1}}}}
\newcommand{\ipder}[3][]{\partial^{#1}{#2}/\partial{#3^{#1}}}
\newcommand{\dder}[3]{\frac{\partial^2{#1}}{\partial{#2}\partial{#3}}}

\newcommand{\wb}[1]{\overline{#1}}
\newcommand{\wt}[1]{\widetilde{#1}}

\def\xssp{\hspace{1pt}}
\def\ssp{\hspace{3pt}}
\def\msp{\hspace{5pt}}
\def\lsp{\hspace{12pt}}

\newcommand{\cA}{\mathcal{A}}
\newcommand{\cB}{\mathcal{B}}
\newcommand{\cC}{\mathcal{C}}
\newcommand{\cD}{\mathcal{D}}
\newcommand{\cE}{\mathcal{E}}
\newcommand{\cF}{\mathcal{F}}
\newcommand{\cG}{\mathcal{G}}
\newcommand{\cH}{\mathcal{H}}
\newcommand{\cI}{\mathcal{I}}
\newcommand{\cJ}{\mathcal{J}}
\newcommand{\cK}{\mathcal{K}}
\newcommand{\cL}{\mathcal{L}}
\newcommand{\cM}{\mathcal{M}}
\newcommand{\cN}{\mathcal{N}}
\newcommand{\cO}{\mathcal{O}}
\newcommand{\cP}{\mathcal{P}}
\newcommand{\cQ}{\mathcal{Q}}
\newcommand{\cR}{\mathcal{R}}
\newcommand{\cS}{\mathcal{S}}
\newcommand{\cT}{\mathcal{T}}
\newcommand{\cU}{\mathcal{U}}
\newcommand{\cV}{\mathcal{V}}
\newcommand{\cW}{\mathcal{W}}
\newcommand{\cX}{\mathcal{X}}
\newcommand{\cY}{\mathcal{Y}}
\newcommand{\cZ}{\mathcal{Z}}

\newcommand{\vA}{\mathbf{A}}
\newcommand{\vB}{\mathbf{B}}
\newcommand{\vC}{\mathbf{C}}
\newcommand{\vD}{\mathbf{D}}
\newcommand{\vE}{\mathbf{E}}
\newcommand{\vF}{\mathbf{F}}
\newcommand{\vG}{\mathbf{G}}
\newcommand{\vH}{\mathbf{H}}
\newcommand{\vI}{\mathbf{I}}
\newcommand{\vJ}{\mathbf{J}}
\newcommand{\vK}{\mathbf{K}}
\newcommand{\vL}{\mathbf{L}}
\newcommand{\vM}{\mathbf{M}}
\newcommand{\vN}{\mathbf{N}}
\newcommand{\vO}{\mathbf{O}}
\newcommand{\vP}{\mathbf{P}}
\newcommand{\vQ}{\mathbf{Q}}
\newcommand{\vR}{\mathbf{R}}
\newcommand{\vS}{\mathbf{S}}
\newcommand{\vT}{\mathbf{T}}
\newcommand{\vU}{\mathbf{U}}
\newcommand{\vV}{\mathbf{V}}
\newcommand{\vW}{\mathbf{W}}
\newcommand{\vX}{\mathbf{X}}
\newcommand{\vY}{\mathbf{Y}}
\newcommand{\vZ}{\mathbf{Z}}

\newcommand{\va}{\mathbf{a}}
\newcommand{\vb}{\mathbf{b}}
\newcommand{\vc}{\mathbf{c}}
\newcommand{\vd}{\mathbf{d}}
\newcommand{\ve}{\mathbf{e}}
\newcommand{\vf}{\mathbf{f}}
\newcommand{\vg}{\mathbf{g}}
\newcommand{\vh}{\mathbf{h}}
\newcommand{\vi}{\mathbf{i}}
\newcommand{\vj}{\mathbf{j}}
\newcommand{\vk}{\mathbf{k}}
\newcommand{\vl}{\mathbf{l}}
\newcommand{\vm}{\mathbf{m}}
\newcommand{\vn}{\mathbf{n}}
\newcommand{\vo}{\mathbf{o}}
\newcommand{\vp}{\mathbf{p}}
\newcommand{\vq}{\mathbf{q}}
\newcommand{\vr}{\mathbf{r}}
\newcommand{\Vs}{\mathbf{s}}
\newcommand{\vt}{\mathbf{t}}
\newcommand{\vu}{\mathbf{u}}
\newcommand{\vv}{\mathbf{v}}
\newcommand{\vw}{\mathbf{w}}
\newcommand{\vx}{\mathbf{x}}
\newcommand{\vy}{\mathbf{y}}
\newcommand{\vz}{\mathbf{z}}

\newcommand{\vone}{\mathbf{1}}
\newcommand{\vzero}{\mathbf{0}}

\newcommand{\valpha}{{\boldsymbol{\alpha}}}
\newcommand{\vbeta}{{\boldsymbol{\beta}}}
\newcommand{\vgamma}{{\boldsymbol{\gamma}}}
\newcommand{\vdelta}{{\boldsymbol{\delta}}}
\newcommand{\vepsilon}{{\boldsymbol{\epsilon}}}
\newcommand{\vzeta}{{\boldsymbol{\zeta}}}
\newcommand{\veta}{{\boldsymbol{\eta}}}
\newcommand{\vtheta}{{\boldsymbol{\theta}}}
\newcommand{\viota}{{\boldsymbol{\iota}}}
\newcommand{\vkappa}{{\boldsymbol{\kappa}}}
\newcommand{\vlambda}{{\boldsymbol{\lambda}}}
\newcommand{\vmu}{{\boldsymbol{\mu}}}
\newcommand{\vnu}{{\boldsymbol{\nu}}}
\newcommand{\vxi}{{\boldsymbol{\xi}}}
\newcommand{\vomikron}{{\boldsymbol{\omikron}}}
\newcommand{\vpi}{{\boldsymbol{\pi}}}
\newcommand{\vrho}{{\boldsymbol{\rho}}}
\newcommand{\vsigma}{{\boldsymbol{\sigma}}}
\newcommand{\vtau}{{\boldsymbol{\tau}}}
\newcommand{\vupsilon}{{\boldsymbol{\upsilon}}}
\newcommand{\vphi}{{\boldsymbol{\phi}}}
\newcommand{\vchi}{{\boldsymbol{\chi}}}
\newcommand{\vpsi}{{\boldsymbol{\psi}}}
\newcommand{\vomega}{{\boldsymbol{\omega}}}

\newcommand{\rLambda}{\mathrm{\Lambda}}
\newcommand{\rSigma}{\mathrm{\Sigma}}

\newcommand{\vLambda}{\bm{\rLambda}}
\newcommand{\vSigma}{\bm{\rSigma}}

% big cdot
\makeatletter
\newcommand*\bdot{\mathpalette\bdot@{.7}}
\newcommand*\bdot@[2]{\mathbin{\vcenter{\hbox{\scalebox{#2}{$\m@th#1\bullet$}}}}}
\makeatother

%--------------------------------------------------------------------
% Add a period to the end of an abbreviation unless there's one
% already, then \xspace.
\makeatletter
\DeclareRobustCommand\onedot{\futurelet\@let@token\@onedot}
\def\@onedot{\ifx\@let@token.\else.\null\fi\xspace}
\def\eg{\emph{e.g}\onedot} \def\Eg{\emph{E.g}\onedot}
\def\ie{\emph{i.e}\onedot} \def\Ie{\emph{I.e}\onedot}
\def\cf{\emph{cf}\onedot} \def\Cf{\emph{C.f}\onedot}
\def\etc{\emph{etc}\onedot} \def\vs{\emph{vs}\onedot}
\def\wrt{w.r.t\onedot} \def\dof{d.o.f\onedot} \def\aka{a.k.a\onedot}
\def\etal{\emph{et al}\onedot}
\makeatother

%% file: tex/defn.tex
\newcommand{\simi}{\operatorname{sim}}
\newcommand{\NN}{\operatorname{NN}}

\newcommand{\sym}{\mathrm{sym}}
\newcommand{\asym}{\mathrm{asym}}
\newcommand{\contr}{\mathrm{contr}}
\newcommand{\triplet}{\mathrm{triplet}}
\newcommand{\MS}{\mathrm{MS}}
\newcommand{\reg}{\mathrm{reg}}
\newcommand{\rkd}{\mathrm{RKD}}
\newcommand{\rank}{\mathrm{DR}}

%% file: tex/defn_exp.tex
\newcommand{\ro}{$\mathcal{R}$Oxf}
\newcommand{\rp}{$\mathcal{R}$Par}
\newcommand{\roxf}{$\mathcal{R}$Oxford5k\xspace}
\newcommand{\rpar}{$\mathcal{R}$Paris6k\xspace}

\newcommand{\vgg}{VGG16~\cite{radenovic2018fine}}
\newcommand{\res}{ResNet101~\cite{radenovic2018fine}}

\newcommand{\lcontrp}{Contr$^+$\xspace}
\newcommand{\lcontr}{Contr~\eq{contr}\xspace}
\newcommand{\ltripl}{Triplet~\eq{triplet}\xspace}
\newcommand{\lms}{MS~\eq{multi}\xspace}
\newcommand{\lreg}{Reg~\eq{reg}\xspace}
\newcommand{\lrkd}{RKD~\eq{rkd}\xspace}
\newcommand{\lrank}{DR~\eq{rank}\xspace}

%% file: tex/intro.tex
\section{Introduction}
\label{sec:intro}

Originating in \emph{metric learning}, loss functions based on pairwise distances or similarities~\cite{HaCL06,WSL+14,OXJS16,WHH+19,Cakir_2019_CVPR} are paramount in representation learning. Their power is most notable in category-level tasks where classes at inference are different than classes at learning, for instance \emph{fine-grained classification} \cite{OXJS16,WHH+19}, \emph{few-shot learning}~\cite{vinyals2016,snell2017} \emph{local descriptor learning}~\cite{HLJ+15} and \emph{instance-level retrieval}~\cite{GARL16,RTC16}. There are different ways to use them without supervision~\cite{Iscen_2018_CVPR,Ye_2019_CVPR,cao2019unsupervised} and indeed, they form the basis for modern \emph{unsupervised representation learning}~\cite{misra2019self,he2019momentum,chen2020simple}.

Powerful representations come traditionally with powerful network models~\cite{HZRS16,huang2017}, which are expensive. The search for resource-efficient architectures has lead to the design of lightweight networks for mobile devices~\cite{howard2017mobilenets,sandler2018mobilenetv2,zhang2018shufflenet}, \emph{neural architecture search}~\cite{pham2018efficient,liu2018darts} and \emph{model scaling}~\cite{tan2019efficientnet}. Training of small networks may be facilitated by \emph{knowledge transfer} from larger networks~\cite{hinton2015}. However, both network design and knowledge transfer are commonly performed on classification tasks, using standard cross-entropy.

Focusing on fine-grained classification and retrieval, several recent methods have extended metric learning loss functions
to allow for knowledge transfer from teacher to student models~\cite{chen2018darkrank,kim2019deep,YYL+19,park2019relational}. However, two questions are in order: (a) since transferring a \emph{representation} from one model to another is inherently a continuous task, can't we just use \emph{regression}? (b) apart from knowledge transfer, is the original metric learning task still relevant and what is a simple way to combine the two?

In this work, we focus on the task of \emph{instance-level image retrieval}~\cite{PCISZ07,RIT+18}, which is at the core of metric learning in the sense of using pairwise distances or similarities. In its most well-known form~\cite{GARL16,RTC16}, the task is supervised, but the supervision is originating from automated data analysis rather than humans. As such, apart from noisy, supervision is often incomplete, in the sense that although class labels per example may exist, not all  pairs of examples of the same class are labeled. Hence, one has to work with pairs rather than examples, unlike \eg \emph{face recognition}~\cite{DGXZ19}.

Our work is motivated by the scenario where a \emph{database (gallery)} of images is represented and indexed according to a large model, while \emph{queries} are captured from mobile devices, where a smaller model is the only option. In such scenario, rather than re-indexing the entire database, it is preferable to adapt different smaller models for different end-user devices. In this case, knowledge transfer from the large (teacher) to the small (student) model is not just helping, but the student should really learn to map inputs to the same representation space. We call this task \emph{asymmetric testing}.

More importantly, even if we consider the standard \emph{symmetric testing} task, where both queries and database examples are represented by the same model at inference, we introduce a novel paradigm of using asymmetric representations \emph{at training}, as a knowledge transfer mechanism. We call this paradigm \emph{asymmetric metric learning}. By representing anchors by the student and positives/negatives by the teacher, one can apply any metric learning loss function. This achieves both metric learning and knowledge transfer, without resorting to a linear combination of two loss functions.

In summary, we make the following contributions:

\begin{itemize}
	\item We study the problem of knowledge transfer from a teacher to a student model for the first time in pair-based metric learning for instance-level image retrieval.
	\item In this context, we study the \emph{asymmetric testing} task, where the database is represented by the teacher and queries by the student.
	\item In both symmetric and asymmetric testing, we systematically evaluate 	different teacher and student models, metric learning loss functions (\autoref{sec:loss-label}) and knowledge transfer loss functions (\autoref{sec:loss-teacher}), serving as a benchmark for future work.
	\item We introduce the \emph{asymmetric metric learning} paradigm, an extremely simple mechanism to combine metric learning with knowledge transfer (\autoref{sec:asym}).
\end{itemize}

%% file: tex/related.tex
\section{Related work}
\label{sec:related}

\paragraph{Metric learning}

Historically, metric learning is about unsupervised learning of embeddings according to a pairwise distances~\cite{Tenenbaum97} or similarities~\cite{scholkopf1998nonlinear,BN03}. Modern deep metric learning is mostly \emph{supervised}, with pair labels specifying a set of \emph{positive} and \emph{negative} examples per \emph{anchor} example~\cite{XJRN03}. Standard loss functions are \emph{contrastive}~\cite{HaCL06} and \emph{triplet}~\cite{XJRN03,WSL+14}, operating on one or two pairs, respectively. \emph{Global} loss functions rather operate on an arbitrary number of pairs~\cite{OXJS16,WHH+19,Cakir_2019_CVPR}, similarly to \emph{learning to rank}~\cite{CQL+07,xia2008listwise}. The large number of potential tuples gives rise to \emph{mining}~\cite{Harwood_2017_ICCV,Wu_2017_ICCV} and \emph{memory}~\cite{wang2019cross,wu2018improving} mechanisms. At the other extreme, extensions of cross-entropy operate on \emph{single examples}~\cite{wang2018cosface,DGXZ19}. We focus on pair-based functions in this work, due to the nature of the ground truth~\cite{RTC16,radenovic2018fine}. \emph{Unsupervised} metric learning is gaining momentum~\cite{Iscen_2018_CVPR,Ye_2019_CVPR,cao2019unsupervised}, but we focus on the supervised case, given that it requires no human effort~\cite{RTC16,GARL16}.

\paragraph{Image retrieval}

Instance-level image retrieval, either using local features~\cite{ToAJ13} or global pooling~\cite{JZ14}, has relied on SIFT descriptors~\cite{Lowe04} for more than a decade. \emph{Convolutional networks} quickly outperformed shallow representations, using different \emph{pooling} mechanisms~\cite{RSMC14,TSJ15} and \emph{fine-tuning} on relevant datasets, initially with cross-entropy on noisy labels from the web~\cite{BSCL14} and then with contrastive~\cite{RTC16} and triplet~\cite{GARL16} loss on labels generated from the visual data alone. While the best performance comes from large networks~\cite{HZRS16,RIT+18}, we focus on \emph{small networks}~\cite{sandler2018mobilenetv2,tan2019efficientnet} for the first time. Our \emph{asymmetric test} scenario is equivalent to that of prior studies~\cite{hu2019towards,shen2020towards}, but with different motivation and settings. \emph{Feature translation}~\cite{hu2019towards} is meant for retrieval system interoperability, so both networks may be large and none is adapted. The recent \emph{backward-compatible training} (BCT)~\cite{shen2020towards} is meant to avoid re-indexing of the database like here, but the new model used for queries is actually more powerful than the old one used for the database, or trained on more data.

\paragraph{Small networks}

While large networks~\cite{HZRS16,huang2017} excel in performance, they are expensive. One solution is to \emph{compress} existing architectures, \eg by quantization~\cite{gong2014compressing} or pruning~\cite{liu2018rethinking}. Another is to manually \emph{design} more efficient networks, \eg by using bottlenecks~\cite{iandola2016squeezenet}, separable convolutions~\cite{howard2017mobilenets}, inverted residuals~\cite{sandler2018mobilenetv2} or point-wise group convolutions ~\cite{zhang2018shufflenet}. MobileNetV2~\cite{sandler2018mobilenetv2} is such a network that we use as a student in this work. More recently, \emph{neural architecture search}~\cite{pham2018efficient,liu2018darts,tan2019mnasnet,howard2019searching} is making this process automatic, although expensive. Alternatively, a small model can first be designed (or learned) and then its architecture \emph{scaled} by adding depth~\cite{HZRS16}, width~\cite{zagoruyko2016wide}, resolution~\cite{huang2019gpipe} or a compound of the above~\cite{tan2019efficientnet}. We use the latter as another student in this work. We show that \emph{pruning}~\cite{wang2020progressive} cannot compete designed or learned architectures.

\paragraph{Knowledge transfer}

Rather than training a small network directly, it is easier to optimize the same small network (\emph{student}) to mimic a larger one (\emph{teacher}), essentially transferring knowledge from the teacher to the student. In classification, this can be done \eg by regression of the logits~\cite{ba2014deep} or by cross-entropy on soft targets, known as \emph{knowledge distillation}~\cite{hinton2015}. BCT~\cite{shen2020towards} fixes the classifier (last layer) of the student to that of the teacher, similarly to~\cite{hoffer2018fix}. Such ideas do not apply in this work, since there is no parametric classifier. \emph{Metric learning} is mostly about pairs rather than individual examples, and indeed recent knowledge transfer methods are based on pairwise distances or similarities. This includes \eg \emph{learning to rank}~\cite{chen2018darkrank} and regression on quantities involving one or more pairs like \emph{distances}~\cite{YYL+19,park2019relational}, \emph{log-ratio of distances}~\cite{kim2019deep}, or \emph{angles}~\cite{park2019relational}. The most general form is \emph{relational knowledge distillation} (RKD)~\cite{park2019relational}. Direct regression on \emph{features} is either not considered or shown inferior~\cite{YYL+19}, but we show it is much more effective than previously thought. We also show that the original metric learning task is still beneficial when training the student and we introduce a very simple mechanism to combine with knowledge transfer.

\paragraph{Asymmetry}

Asymmetric distances or similarities are common in \emph{approximate nearest neighbor search}, where queries may be quantized differently than the database, or not at all~\cite{DoCL08,GoPe11,JeDS11,JJGO11,NSS+13,DaBS14}. In \emph{image retrieval}, there are efforts to reduce the asymmetry of $k$-nearest neighbor relations~\cite{JeHS07}, or use asymmetry to mitigate the effect of quantization~\cite{JeDS10}, or handle partial similarity~\cite{ZhJI13} or alignment~\cite{tolias2017asymmetric}. In \emph{classification}, it is common to use asymmetric image-to-class distances~\cite{BoSI08} or region-to-image matching~\cite{KiGr10}. In \emph{metric learning}, asymmetry has been used in sample weighting~\cite{LiLi15}, different mappings per view~\cite{DBLP:conf/iccv/YuWZ17}, or hard example mining~\cite{XYDZ19}. Asymmetric similarities are used between \emph{cross-modal embeddings}~\cite{frome2013devise,kiros2014unifying,faghri2017vse++}, but not for knowledge transfer. They are also used over the same modality to adjust embeddings to a memory bank~\cite{he2019momentum} or to treat a set of examples as a whole~\cite{vinyals2016}, but again not for knowledge transfer.

%% file: tex/method.tex
\section{Asymmetric metric learning}
\label{sec:method}

\subsection{Preliminaries}
\label{sec:prelim}

Let $X \subset \cX$ be a \emph{training set}, where $\cX$ is an \emph{input space}. Two sources of supervision are considered. The first is a set of \emph{labels}: a subset of all pairs of examples in $X$ is labeled as positive or negative and the remaining are unlabeled. Formally, for each \emph{anchor} $a \in X$, a set $P(a) \subset X$ of \emph{positive} and a set $N(a) \subset X$ of \emph{negative} examples are given. The second is a \emph{teacher} model $g: \cX \to \real^d$, mapping input examples to a \emph{feature (embedding)} space of dimensionality $d$. The objective is to learn the parameters $\theta$ of a \emph{student} model $f_\theta: \cX \to \real^d$, such that anchors are closer to positives than negatives, the teacher and student agree in some sense, or both. When labels are not used, an additional set of examples $U(a)$ may be used for each anchor $a$, \eg a \emph{neighborhood} of $a$ space or the entire set $X \setminus \{a\}$. The teacher is assumed to have been trained on $X$ using labels only.

Training can be formulated as minimizing the \emph{error function}
\begin{align}
	J(X; \theta) \defn \sum_{a \in X} \ell(a; \theta)
\label{eq:error}
\end{align}
with respect to parameters $\theta$ over $X$. There is one loss term per anchor $a \in X$, which however may depend on any other example in $X$; hence, $J$ is not additive in $X$. The \emph{loss function} $\ell$ may depend on the labels or the teacher only, discussed respectively in \autoref{sec:loss-label} and \autoref{sec:loss-teacher}; it may depend on the teacher indirectly via a \emph{similarity function}, as discussed in~\autoref{sec:asym}.

At inference, a \emph{test set} $Z \subset \cX$ and a set of \emph{queries} $Q \subset \cX$ are given, both disjoint from $X$. For each \emph{query} $q \in Q$, a set $P(q) \subset Z$ of \emph{positive} examples is given. \emph{Symmetric testing} is the task of ranking positive examples $P(q)$ before all others in $Z$ by descending similarity to $q$ in the student space, for each query $q \in Q$. \emph{Asymmetric testing} is the same, except that similarities are between queries in the student space and test examples in the teacher space.

%------------------------------------------------------------------------------

\subsection{Asymmetric similarity}
\label{sec:asym}

We use \emph{cosine similarity} in this work: $\simi(\vv, \vv') \defn \inner{\vv, \vv'} / (\norm{\vv} \norm{\vv'})$ for $\vv, \vv' \in \real^d$. The \emph{symmetric similarity} $s_\theta^{\sym}(a, x)$ between an anchor $a \in X$ and a positive or negative example $x \in P(a) \cup N(a)$ is obtained by representing both in the feature space of the student:
\begin{align}
	s_\theta^{\sym}(a, x) \defn \simi(f_\theta(a), f_\theta(x)).
\label{eq:sym}
\end{align}
This is the standard setting in related work in metric learning.

By contrast, we introduce the \emph{asymmetric similarity} $s_\theta^{\asym}(a, x)$, where the anchor $a$ is represented by the student, while positive and negative examples $x$ are represented by the teacher:
\begin{align}
	s_\theta^{\asym}(a, x) \defn \simi(f_\theta(a), g(x)).
\label{eq:asym}
\end{align}
In this setting, $g(x)$ is fixed for all $x \in X$, because the teacher is fixed.

%------------------------------------------------------------------------------

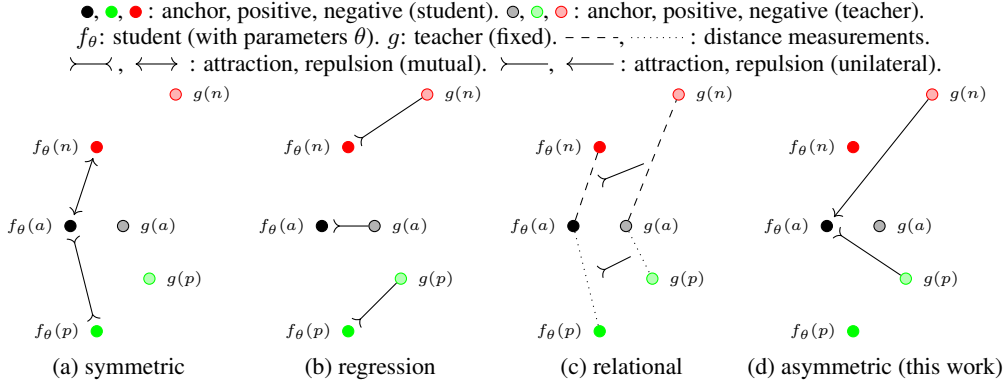
\begin{figure}
\input{tex/idea}
\caption{Metric learning and knowledge transfer. (a) \emph{Symmetric}: Positive (negative) pairs of examples mutually attracted (repulsed) in student space; teacher not used.(b) \emph{Regression} (absolute ML+KD~\cite{YYL+19}): Examples in student space attracted to corresponding examples in teacher space; labels not used. (c) \emph{Relational} (relative ML+KD~\cite{YYL+19} or distance-wise RKD~\cite{park2019relational}): Distances encouraged to be the same in both spaces; labels not used. (d) \emph{Asymmetric} (this work): Anchors in student space attracted to (repulsed from) positives (negatives) in teacher space; both labels and teacher used.}
\label{fig:idea}
\end{figure}

%------------------------------------------------------------------------------

\autoref{fig:idea} illustrates the idea. When used with loss functions discussed in \autoref{sec:loss-label},~\eq{asym} (\autoref{fig:idea}(d)) uses both the labels and the teacher, essentially combining metric learning and knowledge transfer. With the same loss functions,~\eq{sym} (\autoref{fig:idea}(a)) uses the labels only, focusing on metric learning only. Instead, as discussed in \autoref{sec:loss-teacher}, relational distillation~\cite{YYL+19,park2019relational} uses the teacher only, focusing on knowledge transfer only (\autoref{fig:idea}(b,c)). In practice, these other solutions require a linear combination of two error functions for metric learning and knowledge transfer.

%------------------------------------------------------------------------------

\subsection{Loss functions using labels}
\label{sec:loss-label}

When using the labels, we have access to positive and negative examples $P(a)$ and $N(a)$ per anchor $a$. The teacher $g$ may be used in addition to labels or not by using the  asymmetric~\eq{asym} or symmetric~\eq{sym} similarity, respectively. We write either as $s_\theta(a, x)$ below.

\paragraph{Contrastive} The \emph{contrastive} loss~\cite{HaCL06} encourages independently positive examples $p$ to be close to the anchor $a$ and negative examples $n$ farther from $a$ by margin $m$ in the student space:
\begin{align}
	\ell_{\contr}(a; \theta) \defn
		- \sum_{p \in P(a)} s_\theta(a, p)
		+ \sum_{n \in N(a)} [s_\theta(a, n) - m]_+.
\label{eq:contr}
\end{align}

\paragraph{Triplet}

The \emph{triplet} loss~\cite{WSL+14} encourages positive examples $p$ to be closer to the anchor $a$ than negative examples $n$ by margin $m$ in the student space:
\begin{align}
	\ell_{\triplet}(a; \theta) \defn
		\sum_{(p,n) \in L(a)} [s_\theta(a, n) - s_\theta(a, p) + m]_+,
\label{eq:triplet}
\end{align}
where typically $L(a) \defn P(a) \times N(a)$. Positive and negative examples are not used independently: if similarities are ranked correctly, the corresponding loss term is zero.

\paragraph{Multi-similarity} The \emph{multi-similarity} loss~\cite{WHH+19}
treats positives and negatives independently: 
\begin{align}
	\ell_{\MS}(a; \theta) \defn
		\frac{1}{\alpha} \log \left( 1 + \sum_{p \in P(a)} e^{-\alpha(s_\theta(a, p) - m)} \right) +
		\frac{1}{\beta} \log \left( 1 + \sum_{n \in N(a)} e^{\beta(s_\theta(a, n) - m)} \right).
\label{eq:multi}
\end{align}
Here, multiple examples are taken into account together by a nonlinear function: positives (negatives) that are farthest from (nearest to) the anchor receive the greatest relative weight.

%------------------------------------------------------------------------------

\subsection{Loss functions using the teacher only}
\label{sec:loss-teacher}

When not using the labels, the only source of supervision is the teacher model $g$. Symmetric similarity~\eq{sym} is not an option here; we either use use~\eq{asym} or other ways to compare the two models. Given anchor $a$, the loss may depend on $a$ alone, or also the additional examples $U(a)$. We write $S(a, x) \defn \simi(g(a), g(x))$ for the similarity of $a$ and some $x \in U(a)$ in the teacher space.

\paragraph{Regression}

The simplest option is \emph{regression}, encouraging the representations of the same input example $a$ by
the two models to be close by using asymmetric similarity~\eq{asym}:
\begin{align}
	\ell_{\reg}(a; \theta) \defn - s_\theta^{\asym}(a, a) = - \simi(f_\theta(a), g(a)).
\label{eq:reg}
\end{align}
For each anchor, it does not depend on any other example. It is the same as the \emph{absolute} version of \emph{metric learning knowledge distillation} (ML+KD)~\cite{YYL+19} and as contrastive loss~\eq{contr} on asymmetric similarity~\eq{asym} (using only the anchor as a positive for \emph{itself}).

\paragraph{Relational distillation}

Given an anchor $\va$ and one or more other vectors $\vx, \dots$ $ \in \real^d$, \emph{relational knowledge distillation} (RKD)~\cite{park2019relational} is based on a number of relational measurements $\psi(\va, \vx, \dots)$. One such $\psi(\va, \vx, \dots)$ is the \emph{distance} $\norm{\va - \vx}$ for $\vx \in \real^d$. Another is the \emph{angle} $\simi(\va-\vx, \va-\vy)$ formed by $\va, \vx, \vy$, for $\vx, \vy \in \real^d$. The loss is called \emph{distance-wise} and \emph{angle-wise}, respectively. The RKD loss encourages the same measurements
by both models, 
\begin{align}
	\ell_{\rkd}(a; \theta) \defn
		\sum_{(x,\dots) \in U(a)^n}
			r(\psi(f_\theta(a), f_\theta(x),\dots), \psi(g(a), g(x),\dots)),
\label{eq:rkd}
\end{align}
where $n$ is \eg $1$ for distance and $2$ for angle and $r$ is a regression loss, taken as Huber~\cite{park2019relational}. RKD encompasses regression by $\psi$ taken as the identity mapping on the anchor feature alone and $r$ taken as $-\simi$. It also encompasses the \emph{relative} setting of ML+KD~\cite{YYL+19} by $\psi(\va, \vx) \defn \norm{\va - \vx}$ and the \emph{direct match} baseline of DarkRank~\cite{chen2018darkrank} by $\psi(\va, \vx) \defn \norm{\va - \vx}^2$ .

%------------------------------------------------------------------------------

\paragraph{DarkRank}

Let $V(a, x) \defn \{y \in U(a): S(a, y) \le S(a, x) \}$ be the set of examples in $U(a)$ that are mapped farther away from anchor $a$ than $x$ in the teacher space. For each $x \in U(a)$, DarkRank~\cite{chen2018darkrank} encourages those examples to be farther away from $a$ than $x$ in the student space:
\begin{align}
	\ell_{\rank}(a; \theta) \defn
		-\sum_{x \in U(a)} \left(
			s_\theta^{\sym}(a, x) - \log \sum_{y \in V(a, x)} e^{s_\theta^{\sym}(a, y)}
		\right)
\label{eq:rank}
\end{align}
It is an application of the \emph{listwise} loss~\cite{CQL+07,xia2008listwise}, where the ground truth ranking is obtained by the teacher rather than some form of annotation.

%% file: tex/idea.tex
\small
\centering
\setlength\tabcolsep{0pt}

\tikzstyle{idea} = [
	scale=.7,
	font=\tiny,
	anch/.style={black},
	posi/.style={green},
	nega/.style={red},
	tea/.style={pt,fill opacity=.3},
	stu/.style={pt},
	lab/.style={node distance=1pt},
	att-bi/.style={{>[sep=2pt]}-{<[sep=2pt]}},
	rep-bi/.style={{<[sep=2pt]}-{>[sep=2pt]}},
	att/.style={{>[sep=2pt]}-{[sep=2pt]}},
	rep/.style={{<[sep=2pt]}-{[sep=2pt]}},
]

\newcommand{\cell}{
	\node[stu,anch] (sa) at (0,   1)   {};
	\node[stu,posi] (sp) at (0.5,-1)   {};
	\node[stu,nega] (sn) at (0.5, 2.5) {};

	\node[tea,anch] (ta) at (1,   1)   {};
	\node[tea,posi] (tp) at (1.5, 0)   {};
	\node[tea,nega] (tn) at (2,   3.5) {};

	\node[lab,left=of sa] {$f_\theta(a)$};
	\node[lab,left=of sp] {$f_\theta(p)$};
	\node[lab,left=of sn] {$f_\theta(n)$};

	\node[lab,right=of ta] {$g(a)$};
	\node[lab,right=of tp] {$g(p)$};
	\node[lab,right=of tn] {$g(n)$};
}

%------------------------------------------------------------------------------

\extfig{sa}{\tikz[idea] \node[stu,anch]{};},
\extfig{sp}{\tikz[idea] \node[stu,posi]{};},
\extfig{sn}{\tikz[idea] \node[stu,nega]{};}
: anchor, positive, negative (student).
\extfig{ta}{\tikz[idea] \node[tea,anch]{};},
\extfig{tp}{\tikz[idea] \node[tea,posi]{};},
\extfig{tn}{\tikz[idea] \node[tea,nega]{};}
: anchor, positive, negative (teacher).

$f_\theta$: student (with parameters $\theta$).
$g$: teacher (fixed).
\extfig{dash}{\tikz[idea] \draw[dashed] (0,0)--(1,0) (0,-.12);},
\extfig{dott}{\tikz[idea] \draw[dotted] (0,0)--(1,0) (0,-.12);}
: distance measurements.

\extfig{ab}{\tikz[idea] \draw[att-bi] (0,0)--(1,0);},
\extfig{rb}{\tikz[idea] \draw[rep-bi] (0,0)--(1,0);}
: attraction, repulsion (mutual).
\extfig{au}{\tikz[idea] \draw[att] (0,0)--(1,0);},
\extfig{ru}{\tikz[idea] \draw[rep] (0,0)--(1,0);}
: attraction, repulsion (unilateral).

%------------------------------------------------------------------------------

\begin{tabular}{cccc}
\extfig{sym}{
\begin{tikzpicture}[idea]
\cell
\draw
	(sa) edge[att-bi] (sp)
	(sa) edge[rep-bi] (sn);
\end{tikzpicture}
}
&
\extfig{reg}{
\begin{tikzpicture}[idea]
\cell
\draw[att]
	(sa) edge (ta)
	(sp) edge (tp)
	(sn) edge (tn);
\end{tikzpicture}
}
&
\extfig{rel}{
\begin{tikzpicture}[idea]
\cell
\draw[dotted]
	(sa)--(sp) node[midway](spm){}
	(ta)--(tp) node[midway](tpm){};
\draw[dashed]
	(sa)--(sn) node[midway](snm){}
	(ta)--(tn) node[midway](tnm){};
\draw[att]
	(spm) edge (tpm)
	(snm) edge (tnm);
\end{tikzpicture}
}
&
\extfig{asym}{
\begin{tikzpicture}[idea]
\cell
\draw
	(sa) edge[att] (tp)
	(sa) edge[rep] (tn);
\end{tikzpicture}
}
\\
(a) symmetric &
(b) regression &
(c) relational &
(d) asymmetric (this work)
\end{tabular}

%% file: tex/exp_setup.tex
\section{Experiments}
\label{sec:exp}

\subsection{Setup}
\label{sec:setup}

\paragraph{Datasets}

We use the \emph{SfM} dataset~\cite{radenovic2018fine} for training, containing 133k images for training and 30k images for validation. We use the \emph{revisited $\mathcal{R}$Oxford5k} and \emph{$\mathcal{R}$Paris6k} datasets~\cite{RIT+18} for testing, each having 70 query images. All datasets depict particular architectural landmarks under very diverse viewing conditions. We follow the standard evaluation protocol, using the \textit{medium} and \textit{hard} settings~\cite{RIT+18}. We report \emph{mean average precision} (mAP), including \emph{mean precision at 10} (mP@10) in the supplementary material. Comparisons are based on mAP. To compare with pruning~\cite{wang2020progressive}, we also use the original \emph{Oxford5k}~\cite{PCISZ07} and \emph{Paris6k}~\cite{PCS+08} datasets, reporting mAP only.

\paragraph{Networks}

All models are pre-trained for classification on ImageNet~\cite{russakovsky2015imagenet} and then fine-tuned for image retrieval on SfM, following the setup of the same work.
We use VGG-16~\cite{simonyan2014very} and ResNet101~\cite{HZRS16} as \emph{teacher} models with the feature dimensionality $d$ of 512 and 2048, respectively. We use MobileNetV2~\cite{sandler2018mobilenetv2} and EfficientNet-B3~\cite{tan2019efficientnet} as \emph{student} networks, removing any fully connected layers and stacking one $1\times1$ convolutional layer to match the dimensionality of the teacher. All networks use \emph{generalized mean-pooling} (GeM)~\cite{radenovic2018fine} on the last convolutional feature map.

\paragraph{Implementation details}

The image resolution is limited to $362\times 362$ at training (fine-tuning). At testing, a \emph{multi-scale} representation is used, with initial resolution of $1024\times1024$ and scale factors of 1, $\frac{1}{\sqrt{2}}$ and $\frac{1}{2}$. The representation is pooled by GeM over the features of the three scaled inputs. We use \emph{supervised whitening}, trained on the same SfM dataset~\cite{radenovic2018fine}. In \emph{asymmetric testing}, whitening is learned in the teacher space. Our implementation is based on the official code of~\cite{radenovic2018fine} in PyTorch,\footnote{\url{https://github.com/filipradenovic/cnnimageretrieval-pytorch}} as well as~\cite{WHH+19,park2019relational,chen2018darkrank}. Teacher models are taken from~\cite{radenovic2018fine}.

\paragraph{Training and hyper-parameters}  

We follow the training setup of~\cite{radenovic2018fine} for loss functions that use labels. We use the validation set to determine the hyperparameter values and the best model. We train all models using the SGD with learning rate decay of 0.99 per epoch. \emph{Symmetric training}~\eq{sym} takes place for 100 epochs or until convergence based on the validation set. For \emph{asymmetric training}~\eq{asym}, this is extended to 300 epochs. Each epoch consists of 2000 tuples. A mini-batch has 10 tuples, each composed of 1 anchor, 1 corresponding positive and 5 negatives. For unsupervised losses we create tuples of the same overall size. We use weight decay of $10^{-6}$ in each experiment.

\paragraph{Loss functions} 

For \emph{contrastive} loss~\eq{contr}, we set the margin $m=0.7$ and the initial learning rate $\eta$ to $10^{-5}$ and $10^{-3}$ for symmetric and asymmetric training, respectively. For \emph{triplet}~\eq{triplet}, we set $m=0.1$ and $\eta=10^{-8}$ . For \emph{multi-similarity}~\eq{multi} we set $m=0.6$, $\alpha=1$, $\beta=1$ and $\eta=10^{-8}$ for all setups. For \emph{regression}~\eq{reg}, we set $\eta=10^{-3}$. We use the DA variant or RKD~\cite{park2019relational}~\eq{rkd}, with the angle-wise and distance-wise loss weighted by a factor of 2 and 1 respectively, and $\eta=10^{-2}$. For DarkRank (DR)~\eq{rank}, we set $\eta=10^{-6}$ for the VGG16 teacher; for ResNet101, $\eta=10^{-5}$ for MobileNetV2 and $\eta=10^{-7}$ for EfficientNet-B3. We do not discriminate between the student training being \emph{supervised} or not, since labels are already used for teacher training.

\paragraph{Mining}

When \emph{using labels}, we use \emph{hard negative mining} as a default, following~\cite{radenovic2018fine}. Negatives are mined each epoch from a random subset of 22k images of the training set. The negatives closest to the anchor (according to~\eq{sym} or~\eq{asym}, depending on the setting) are selected. There is no mining for positives, because there are only few (1-2) positives per anchor. When \emph{not using labels}, we draw additional examples uniformly at \emph{random} as a default. There is no mining for regression.

%% file: tex/exp_results.tex
\subsection{Results}
\label{sec:results}

%------------------------------------------------------------------------------
\begin{table}
\centering
\scriptsize %\footnotesize
\setlength\tabcolsep{1.5pt}
\begin{tabular}{llcccccccccccccccc}
	\toprule
	\mr{3}{\Th{Student}} & \mr{3}{\Th{Teacher}} && \mr{3}{\Th{Lab}} & \mr{3}{\Th{Loss}} & \mr{3}{\Th{Self}} & \mr{3}{\Th{Pos}} & \mr{3}{\Th{Neg}} & \mr{3}{\Th{Mining}} && \mc{4}{\Th{Symmetric Testing}} & \mc{4}{\Th{Asymmetric Testing}} \\
	&&&&&&&&& \Th{Asym} & \mc{2}{\Th{Medium}} & \mc{2}{\Th{Hard}} & \mc{2}{\Th{Medium}} & \mc{2}{\Th{Hard}} \\
	&&&&&&&&&& {\ro} & {\rp} & {\ro} & {\rp} & {\ro} & {\rp} & {\ro} & {\rp} \\ \midrule
	\mr{5}{MobileNetV2}     & \mr{5}{VGG16}     && \ch & \lcontr &     & \ch & \ch & hard & \ch & \tb{57.3} & 67.1      & 31.1      & 41.3      & 38.3      & 49.8      & 18.4      & 23.8      \\
	                        &                   && \ch & \lcontr & \ch & \ch & \ch & hard & \ch & \tb{57.3} & \tb{68.4} & \tb{31.5} & \tb{42.2} & 42.9      & 55.9      & 22.6      & 31.4      \\
	                        &                   && \ch & \lcontr &     & \ch &     & hard & \ch & 55.9      & 66.7      & 31.1      & 40.6      & 34.1      & 47.3      & 17.0      & 24.5      \\
	                        &                   && \ch & \lcontr & \ch & \ch &     & hard & \ch & 55.5      & 67.0      & 30.4      & 40.9      & 38.2      & 52.2      & 15.3      & 28.9      \\ \cmidrule{4-18}
	                        &                   &&     & \lreg   & \ch &     &     & --   & \ch & 53.3      & 67.5      & 28.9      & 40.9      & \tb{48.0} & \tb{57.9} & \tb{26.5} & \tb{32.6} \\
	\bottomrule
\end{tabular}
\vspace{3pt}
\caption{\emph{Contrastive--regression ablation}. Symmetric and asymmetric testing mAP on \roxf and \rpar~\cite{RIT+18}. \Th{Lab}: using labels in student model training. \Th{Pos}, \Th{Neg}: Using positives, negatives. \Th{Self}: Using anchor (by teacher) as positive for itself (by student). \Th{Asym}: Using asymmetric similarity~\eq{asym} at training. The second row is an option that we call \lcontrp. GeM pooling and learned whitening~\cite{radenovic2018fine} used in all cases.}
\label{tab:sym_test_ablation}
\end{table}
%------------------------------------------------------------------------------

\paragraph{Contrastive--regression ablation}

As will be shown in the following results, contrastive loss and regression turn out be most effective in general. Moreover, by comparing~\eq{contr} with~\eq{reg}, contrastive with asymmetric similarity~\eq{asym} encompasses regression by setting each anchor as a positive for itself, without any other positive or negatives. To better understand the relation between these two loss functions, we perform an ablation study where we investigate versions of contrastive on~\eq{asym} having negatives, or not, and the anchor itself as positive, or not. The results are shown in \autoref{tab:sym_test_ablation} for VGG16$\to$MobileNetV2. In symmetric testing, it turns out that the best combination is having both negatives and the anchor itself. The same happens in almost all cases for other teacher and student models, as shown in the supplementary material. We denote this combination as \lcontrp and we include it in subsequent results.
Asymmetric testing is much more challenging. The best is regression in this case, but \lcontrp is still the second best.

%------------------------------------------------------------------------------
\begin{table}
\centering
\scriptsize %\footnotesize
\setlength\tabcolsep{1.5pt}
\begin{tabular}{lcccccccccccccccc}
	\toprule
	\mr{3}{\Th{Student}} && \mr{3}{$d$} & \mr{3}{\Th{Teacher}} && \mr{3}{\Th{Lab}} & \mr{3}{\Th{Loss}} & \mr{3}{\Th{Mining}} && \mc{4}{\Th{Symmetric Testing}} & \mc{4}{\Th{Asymmetric Testing}} \\
	&&&&&&&& \Th{Asym} & \mc{2}{\Th{Medium}} & \mc{2}{\Th{Hard}} & \mc{2}{\Th{Medium}} & \mc{2}{\Th{Hard}} \\
	&&&&&&&&& {\ro} & {\rp} & {\ro} & {\rp} & {\ro} & {\rp} & {\ro} & {\rp} \\ \midrule
	\vgg                     && 512          &                   && \ch & \lcontr  & hard   &     & 60.9      & 69.3      & 32.9      & 44.2      &           &           &           &           \\
	\res                     && 2048         &                   && \ch & \lcontr  & hard   &     & 65.4      & 76.7      & 40.1      & 55.2      &           &           &           &           \\
	\mr{2}{MobileNetV2}      && 512          &                   && \ch & \lcontr  & hard   &     & 53.6      & 66.4      & 28.8      & 39.7      &           &           &           &           \\
	                         && 2048         &                   && \ch & \lcontr  & hard   &     & 56.1      & 68.5      & 30.3      & 42.0      &           &           &           &           \\
	\mr{2}{EfficientNet-B3}  && 512          &                   && \ch & \lcontr  & hard   &     & 53.8      & 70.9      & 26.2      & 46.0      &           &           &           &           \\
	                         && 2048         &                   && \ch & \lcontr  & hard   &     & 59.6      & 75.1      & 33.3      & 51.9      &           &           &           &           \\ \midrule
	\mr{16}{MobileNetV2}     && \mr{8}{512}  & \mr{8}{VGG16}     && \ch & \lcontrp & hard   & \ch & \tb{57.3} & \tb{68.4} & \tb{31.5} & \tb{42.2} & 42.9      & 55.9      & 22.6      & 31.4      \\
	                         &&              &                   && \ch & \lcontr  & hard   & \ch & \tb{57.3} & 67.1      & 31.1      & 41.3      & 38.3      & 49.8      & 18.4      & 23.8      \\
	                         &&              &                   && \ch & \ltripl  & hard   & \ch & 37.0      & 62.4      & 11.8      & 36.0      &  1.8      &  4.3      &  0.7      &  2.8      \\
	                         &&              &                   && \ch & \lms     & hard   & \ch & 36.8      & 62.8      & 11.5      & 36.5      &  1.9      &  4.3      &  0.8      &  2.7      \\ \cmidrule{5-17}
	                         &&              &                   &&     & \lreg    & --     & \ch & 53.3      & 67.5      & 28.9      & 40.9      & \tb{48.0} & \tb{57.9} & \tb{26.5} & \tb{32.6} \\
	                         &&              &                   &&     & \lrkd    & random &     & 46.2      & 64.3      & 21.8      & 37.6      &  2.0      &  4.1      &  0.8      &  2.6      \\
	                         &&              &                   &&     & \lrank   & random &     & 45.2      & 60.6      & 24.6      & 33.1      &  1.7      &  3.8      &  0.7      &  2.4      \\ \cmidrule{3-17}
	                         && \mr{8}{2048} & \mr{8}{ResNet101} && \ch & \lcontrp & hard   & \ch & \tb{63.2} & \tb{75.0} & \tb{37.9} & \tb{52.0} & 47.1      & 61.5      & 21.8      & 37.7      \\
	                         &&              &                   && \ch & \lcontr  & hard   & \ch & 60.8      & 72.1      & 36.1      & 47.6      & 32.3      & 51.5      &  9.6      & 28.2      \\
	                         &&              &                   && \ch & \ltripl  & hard   & \ch & 45.5      & 68.0      & 19.6      & 43.4      &  1.3      &  3.7      &  0.7      &  2.4      \\
	                         &&              &                   && \ch & \lms     & hard   & \ch & 44.5      & 68.1      & 17.9      & 43.2      &  1.4      &  3.6      &  0.7      &  2.3      \\ \cmidrule{5-17}
	                         &&              &                   &&     & \lreg    & --     & \ch & 59.8      & 73.1      & 35.7      & 49.5      & \tb{49.2} & \tb{65.0} & \tb{23.3} & \tb{40.7} \\
	                         &&              &                   &&     & \lrkd    & random &     & 56.1      & 69.8      & 31.8      & 44.2      &  1.6      &  4.1      &  0.8      &  2.5      \\
	                         &&              &                   &&     & \lrank   & random &     & 48.3      & 58.0      & 23.3      & 31.5      &  1.3      &  4.1      &  0.6      &  2.7      \\ \midrule
	\mr{16}{EfficientNet-B3} && \mr{8}{512}  & \mr{8}{VGG16}     && \ch & \lcontrp & hard   & \ch & \tb{56.9} & 69.0      & 31.1      & 43.5      & 44.7      & 58.0      & 23.9      & 32.4      \\
	                         &&              &                   && \ch & \lcontr  & hard   & \ch & 56.8      & \tb{70.4} & \tb{31.2} & \tb{45.4} & 43.8      & 24.9      & 23.0      &  6.1      \\
	                         &&              &                   && \ch & \ltripl  & hard   & \ch & 33.7      & 64.6      & 8.0       & 40.3      &  1.4      &  4.0      &  0.6      &  2.5      \\
	                         &&              &                   && \ch & \lms     & hard   & \ch & 33.9      & 64.9      &  8.1      & 40.6      &  1.4      &  3.9      &  0.6      &  2.5      \\ \cmidrule{5-17}
	                         &&              &                   &&     & \lreg    & --     & \ch & 55.0      & 69.4      & 27.1      & 44.5      & \tb{49.4} & \tb{58.2} & \tb{26.0} & \tb{33.0} \\
	                         &&              &                   &&     & \lrkd    & random &     & 51.6      & 67.6      & 26.2      & 41.7      &  1.3      &  3.8      &  0.6      &  2.5      \\
	                         &&              &                   &&     & \lrank   & random &     & 34.7      & 61.1      &  8.5      & 35.2      &  1.4      &  3.9      &  0.6      &  2.4      \\ \cmidrule{3-17}
	                         && \mr{8}{2048} & \mr{8}{ResNet101} && \ch & \lcontrp & hard   & \ch & \tb{66.8} & 77.1      & \tb{42.5} & \tb{55.5} & 45.2      & 63.7      & 19.6      & 40.9      \\
	                         &&              &                   && \ch & \lcontr  & hard   & \ch & 66.3      & \tb{77.4} & 41.3      & \tb{55.5} & 37.4      & 57.4      & 10.9      & 33.7      \\
	                         &&              &                   && \ch & \ltripl  & hard   & \ch & 39.5      & 69.4      & 11.6      & 45.8      &  1.5      &  4.0      &  0.7      &  2.5      \\
	                         &&              &                   && \ch & \lms     & hard   & \ch & 39.9      & 69.7      & 11.7      & 46.2      &  1.5      &  4.0      &  0.7      &  2.4      \\ \cmidrule{5-17}
	                         &&              &                   &&     & \lreg    & --     & \ch & 64.9      & 74.4      & 40.5      & 52.4      & \tb{52.9} & \tb{65.2} & \tb{27.8} & \tb{42.4} \\
	                         &&              &                   &&     & \lrkd    & random &     & 56.3      & 73.0      & 30.5      & 46.4      &  1.6      &  3.8      &  0.7      &  2.4      \\
	                         &&              &                   &&     & \lrank   & random &     & 40.3      & 69.9      & 11.8      & 46.4      &  1.5      &  4.0      &  0.7      &  2.5      \\
	\bottomrule
\end{tabular}
\vspace{3pt}
\caption{\emph{Symmetric and asymmetric testing} mAP on \roxf and \rpar~\cite{RIT+18}. \Th{Lab}: using labels in student model training. \Th{Asym}: Using asymmetric similarity~\eq{asym} at training (our work). \lcontrp is defined in \autoref{tab:sym_test_ablation}. Best result highlighted per teacher-student pair. GeM pooling and learned whitening~\cite{radenovic2018fine} used in all cases.}
\label{tab:sym_asym_test_full}
\end{table}
%------------------------------------------------------------------------------

\paragraph{Symmetric testing}

According to the left part of \autoref{tab:sym_asym_test_full}, \lcontrp works best on MobileNetV2, while on EfficientNet, either contrastive and \lcontrp works best, with the two options having little difference. The difference to other loss functions \emph{using labels} is large, reaching 20\% or even 30\% on \roxf. Triplet is known to be inferior to contrastive~\cite{radenovic2018fine}, but the difference is more pronounced in our knowledge transfer setting. This result is particularly surprising for multi-similarity, which is state of the art in fine-grained classification~\cite{WHH+19}. Also surprisingly, regression works best among loss functions \emph{not using labels}, including recent knowledge transfer methods RKD~\cite{park2019relational} and DarkRank~\cite{chen2018darkrank}. It is second or third best in all cases. This finding is contrary to~\cite{YYL+19}, where the regression baseline is found inferior. DarkRank is inferior to RKD, in agreement with~\cite{park2019relational}.

The superiority of contrastive or \lcontrp over regression confirms that, by using our asymmetric similarity, the original metric learning task is still beneficial. Unlike knowledge distillation on classification tasks, knowledge transfer alone is not the best option. Focusing on the best results (contrastive or \lcontrp), we confirm that, with just one exception (VGG16$\to$EfficientNet on \rpar hard), \emph{knowledge transfer always helps} compared to training
without the teacher, using the same $d$. The gain is more pronounced, reaching 7-10\% on ResNet101$\to$MobileNetV2, when the teacher is stronger and the student is weaker (the exception corresponds to the weakest teacher and strongest student). MobileNetV2 performs only 2-3\% below its teacher. Remarkably, EfficientNet \emph{outperforms its teacher}: this happens on \rpar for VGG16 and on all settings for ResNet101.

\paragraph{Asymmetric testing}

Here, similarities are asymmetric at testing, with the database being represented by the teacher and queries by the student.
According to the right part of \autoref{tab:sym_asym_test_full}, \emph{regression is the clear winner} in this case. This is contrary to~\cite{shen2020towards}, where regression fails. In a sense, this can be expected since the student should learn to map images to features exactly like the teacher. \lcontrp is clearly the second best, with the differences varying between 1-2\% on \rpar and up to 8\% on \roxf. Contrastive is the third, with a further loss of roughly 5-10\% or more. Knowledge transfer of weak information like relations or ranking fails completely in this case, which is totally expected. What is unexpected is that triplet and multi-similarity fail too.

When compared with symmetric testing using the corresponding teacher alone, the loss of using the student on queries is 8-16\%. There is no substantial difference in this behavior between MobileNetV2 and EfficientNet. Asymmetric testing is considerably more challenging than symmetric. The closest work in terms of asymmetric testing is \emph{feature translation}~\cite{hu2019towards}, where a shallow translator is learned instead of fine-tuning the student end-to-end. This approach performs poorly, with up to 40\% mAP loss. Students are still large networks, so there is no computational gain.

%------------------------------------------------------------------------------
\begin{table}
\centering
\scriptsize %\footnotesize
\setlength\tabcolsep{1.5pt}
\begin{tabular}{cccccccc}
	\toprule
	\mr{2}{\Th{Student}} & \mr{2}{\Th{\%FLOPS}} & \mr{2}{\Th{\%Param}} & \mr{2}{\Th{Teacher}} & \mr{2}{\Th{Lab}} & \mr{2}{\Th{Loss}} & \mc{2}{mAP}  \\
	&&&&&& Oxf & Par \\ \midrule
	VGG16~\cite{radenovic2018fine}        & 100          & 100           &                     & \ch & \lcontr  & 82.45      & 81.37      \\
	VGG16-PLFP~\cite{wang2020progressive} & 57.37        & 61.05         &                     & \ch & \lcontr  & \tb{76.20} & 73.18      \\ \midrule
	\mr{4}{MobileNetV2}                   & \mr{2}{2.44} & \mr{2}{19.58} & \mr{2}{VGG16}       & \ch & \lcontr  & 74.14      & 78.26      \\
	                                      &              &               &                     &     & \lreg    & 66.58      & 74.45      \\ \cmidrule{2-8}
	                                      & \mr{2}{3.14} & \mr{2}{32.97} & \mr{2}{ResNet101}   & \ch & \lcontr  & 75.30      & \tb{83.23} \\
	                                      &              &               &                     &     & \lreg    & 63.57      & 76.96      \\ 
 	\bottomrule
\end{tabular}
\vspace{3pt}
\caption{\emph{Symmetric testing} mAP on Paris6k and Oxford5k. FLOPS and parameters relative to VGG16. \Th{Lab}: using labels in student model training. Using asymmetric similarity~\eq{asym} in all teacher-student settings. GeM pooling~\cite{radenovic2018fine} used in all cases but \emph{not} learned whitening.}
\label{tab:sym_test_old}
\end{table}
%------------------------------------------------------------------------------

\paragraph{Results on Paris6k and Oxford5k}

We consider this experiment primarily for comparison with \emph{progressive local filter pruning} (PLFP)~\cite{wang2020progressive}, which performs symmetric testing with a pruned version of VGG16. For the sake of comparison, there is no whitening in this case. According to \autoref{tab:sym_test_old}, MobileNetV2 has substantially lower FLOPS and parameters than the pruned VGG16, yet with either teacher it performs better on Paris and nearly the same on Oxford. EfficientNet, reported in the supplementary material, performs best, although it has more parameters.

%% file: tex/conclusion.tex
\section{Conclusions}
\label{sec:conclusion}

There are certain unexpected or surprising findings in this work. First, regression is particularly effective in knowledge transfer. It appears that the more the constraints on student mappings (\eg ranking $\to$ distance/angle relations $\to$ positions), the better the performance in standard symmetric testing. Second, the standard contrastive loss is particularly effective with asymmetric similarity at training, outperforming by a large margin state of the art methods like multi-similarity. A straightforward combination with regression---treating the anchor itself as positive---performs best on symmetric testing. In the new asymmetric testing task, regression is unsurprisingly a winner.

We have shown that using the original metric learning task while transferring knowledge is still beneficial in symmetric testing. It remains to be investigated whether the same can happen in asymmetric testing. We consider the same dataset in teacher and student training, so the latter is as supervised as the former. An interesting extension would be to consider a different, unlabeled dataset in student training. This would be a semi-supervised solution, like \emph{data distillation}~\cite{RDG+18}.

%% file: tex/appendix.tex
%\title{Supplementary Material of \\ ``Asymmetric metric learning for knowledge transfer''}
%\maketitle

%\setcounter{page}{1}
%\linenumbers[1]

\appendix

\section{Example of asymmetric hard example mining}
\label{sec:mining}

\begin{figure}
\begingroup
\setlength{\tabcolsep}{2pt}
\begin{tabular}{cccccccc}
\Th{Epoch} & $a$ & \hspace{10pt} & \mc{5}{$N(a)$}\\
\raisebox{3.5\height}{0} & \figr[.135]{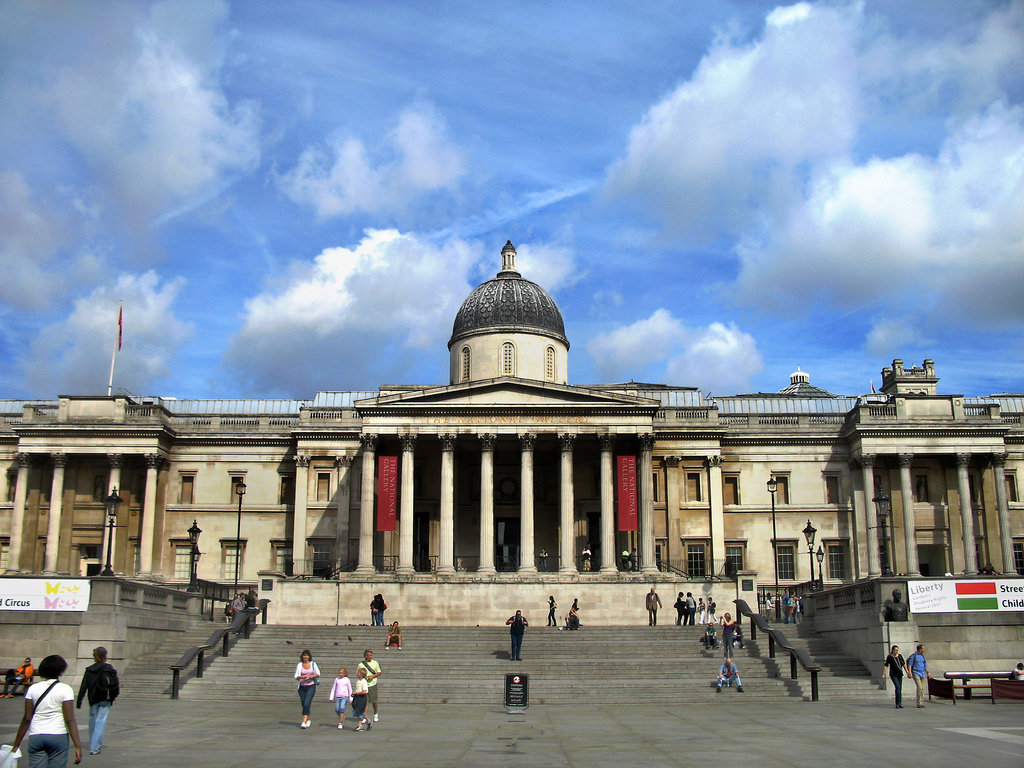} & & \figr[.135]{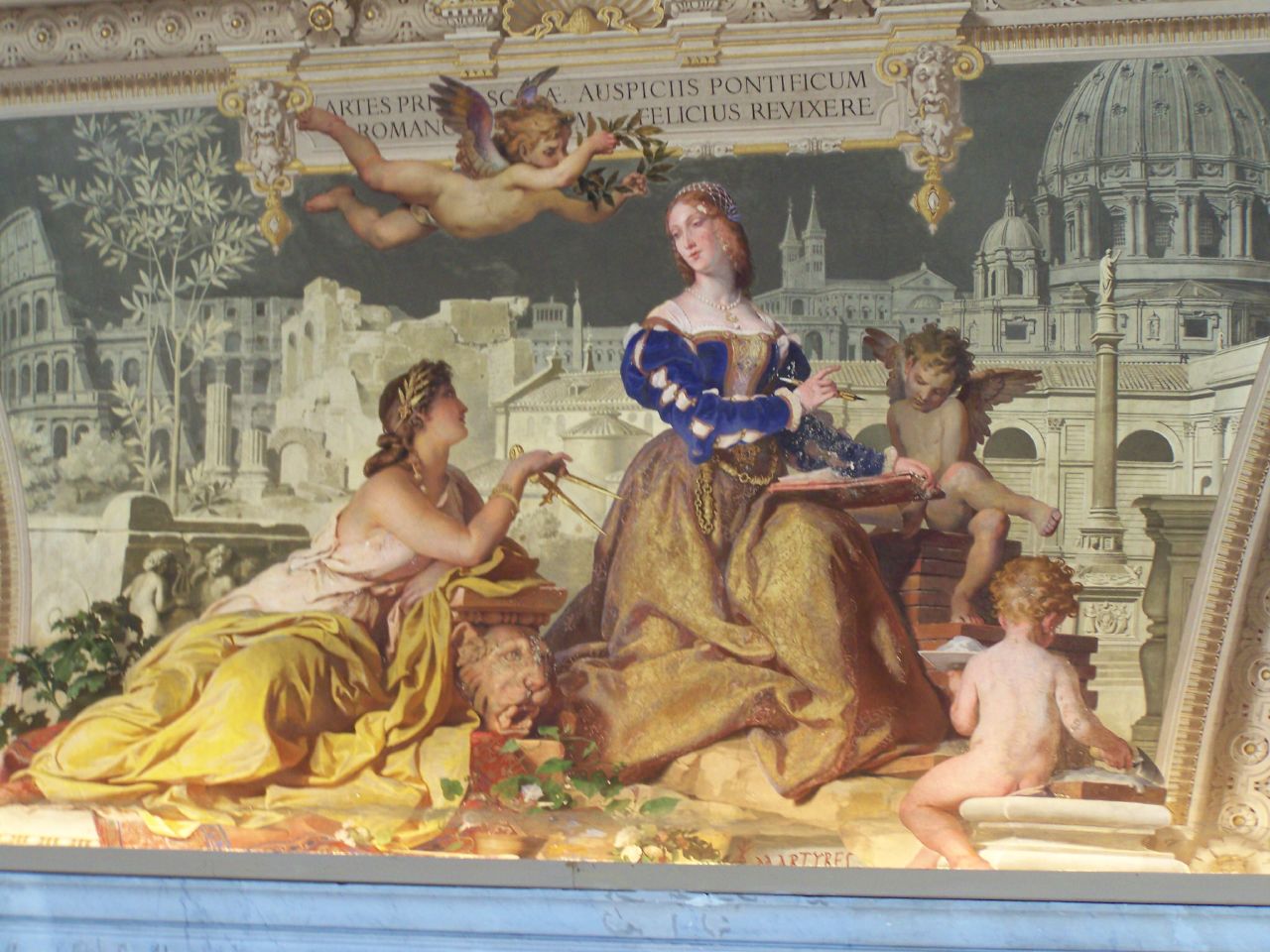} &\figr[.135]{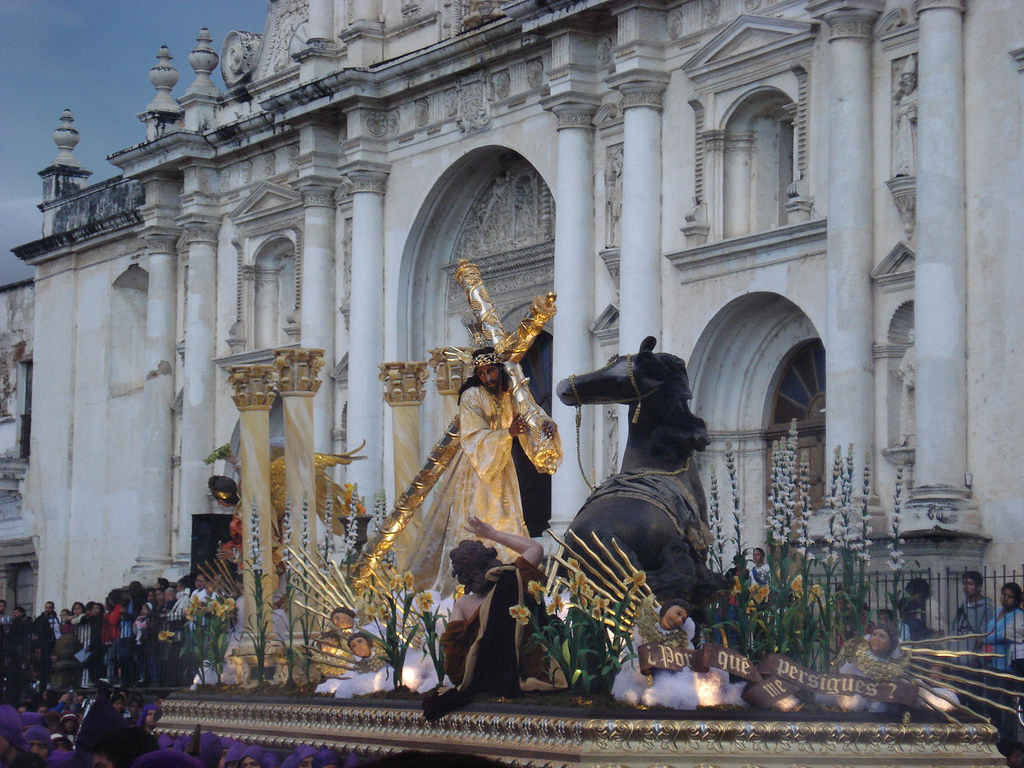} & \figr[.135]{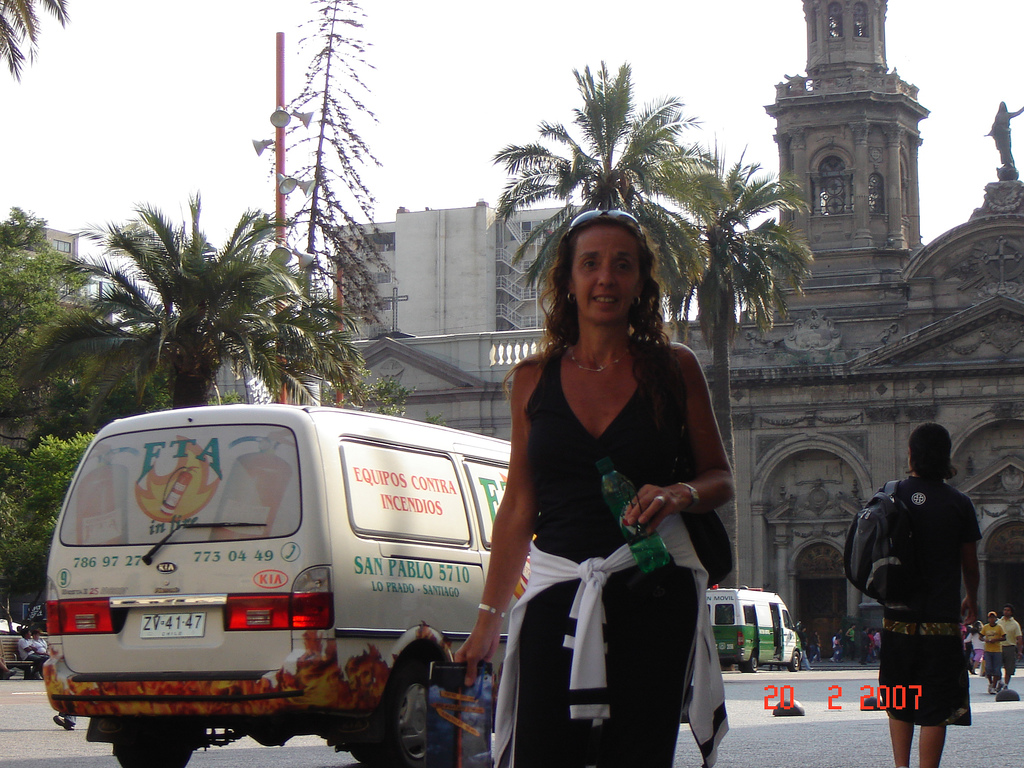} &\figr[.135]{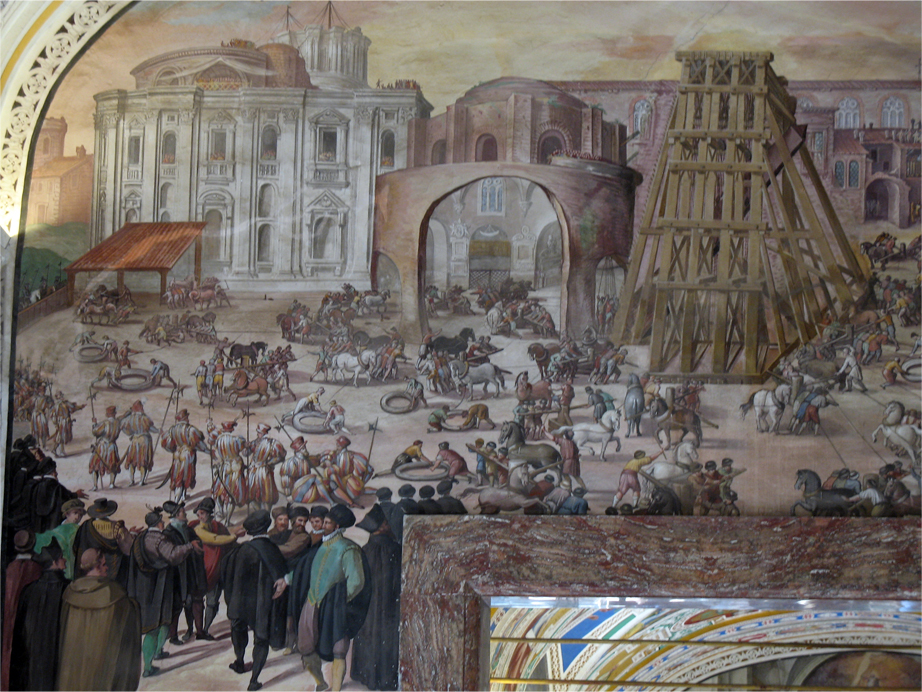} & \figr[.135]{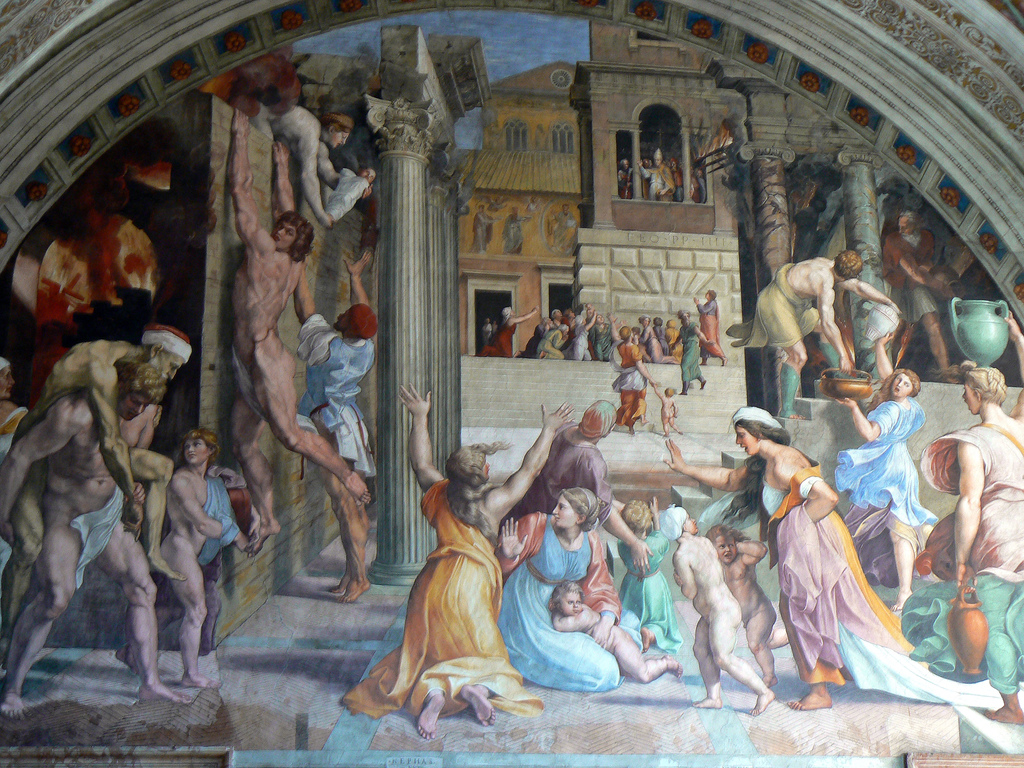} \\
\raisebox{3.5\height}{5} & \figr[.135]{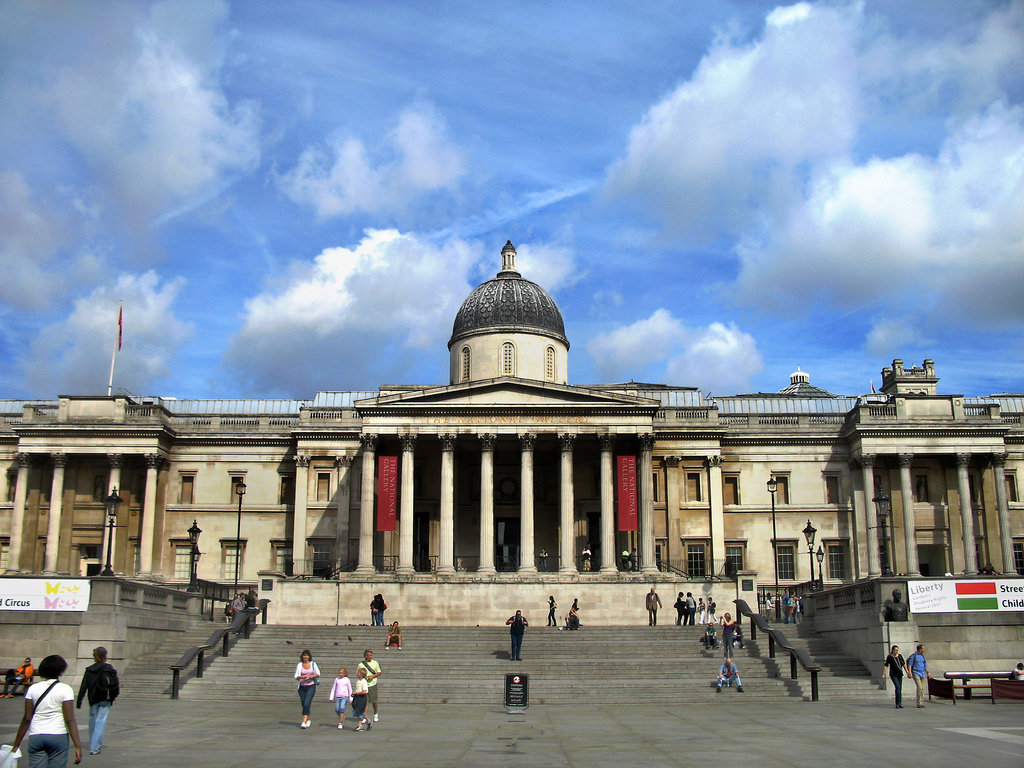} & & \figr[.135]{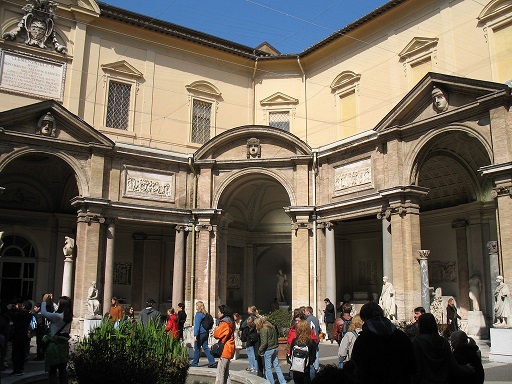} &\figr[.135]{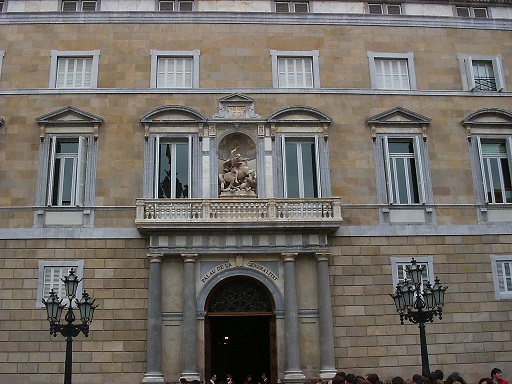} & \figr[.135]{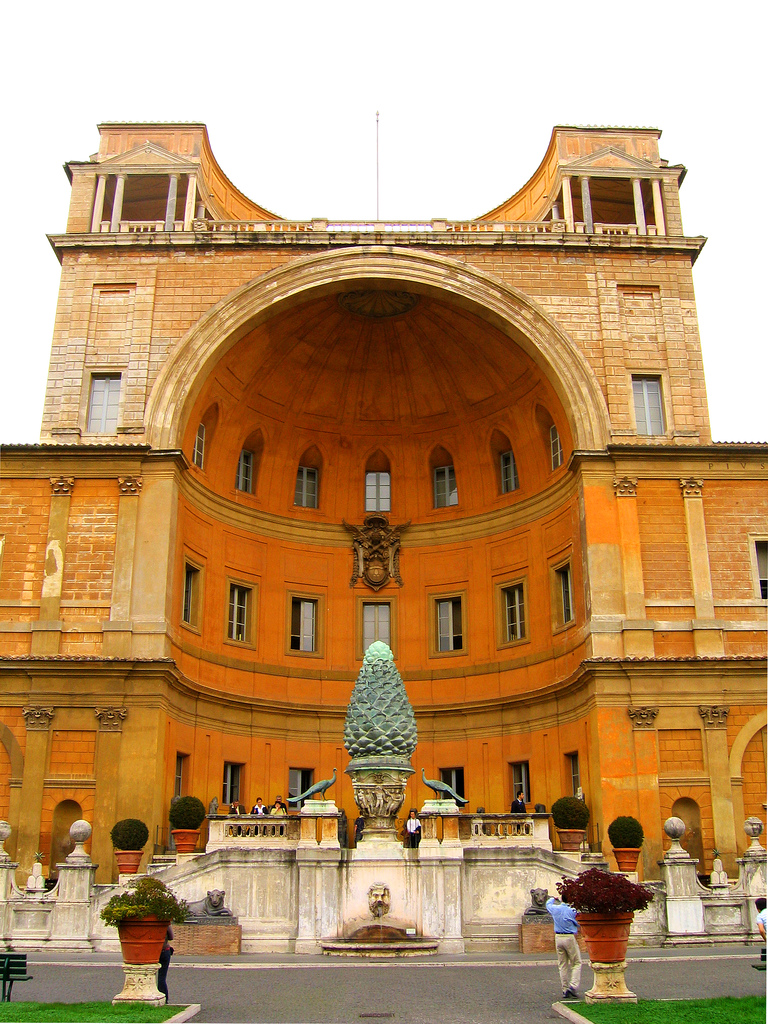} &\figr[.135]{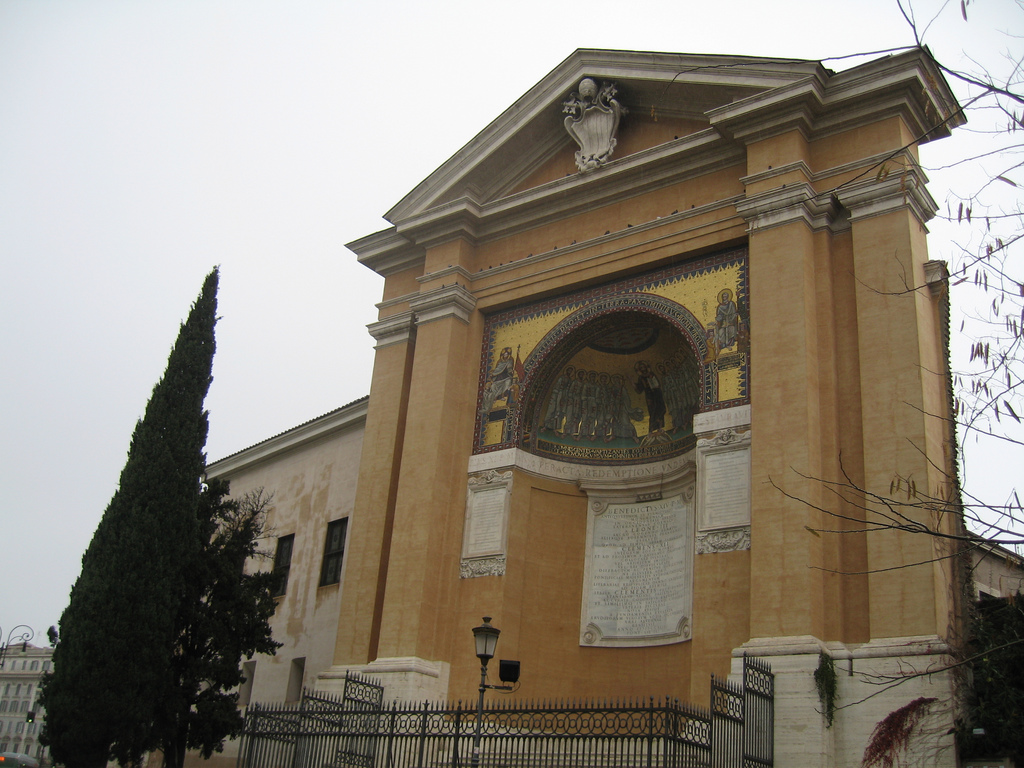} & \figr[.135]{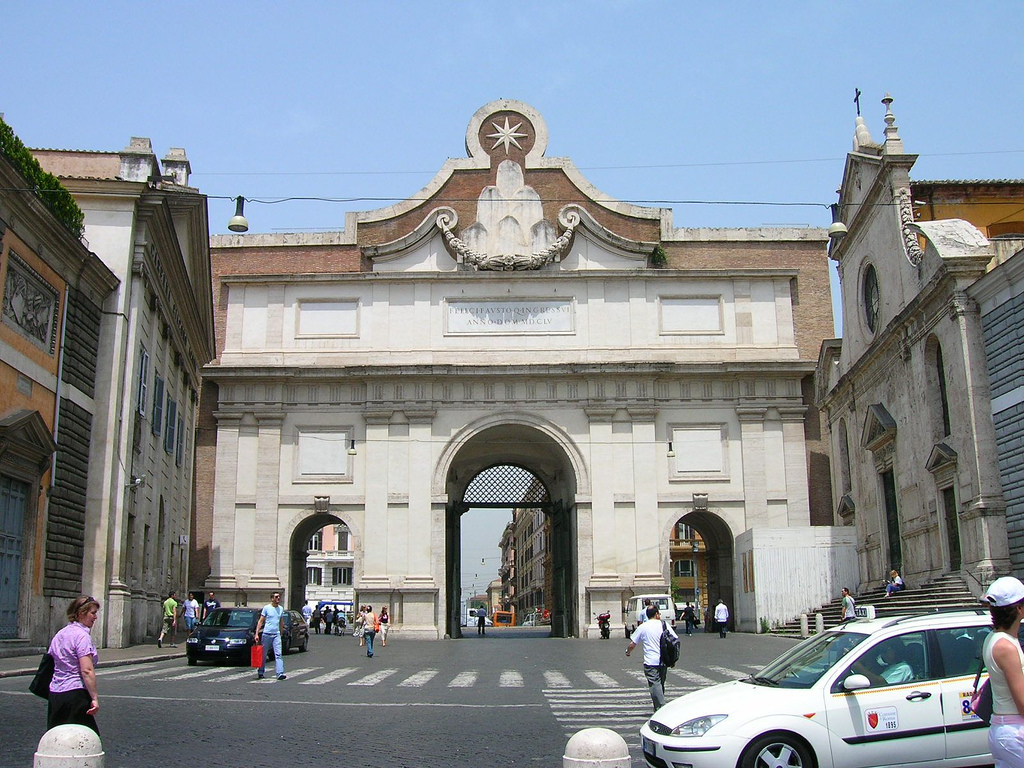} \\
\raisebox{3.5\height}{10} & \figr[.135]{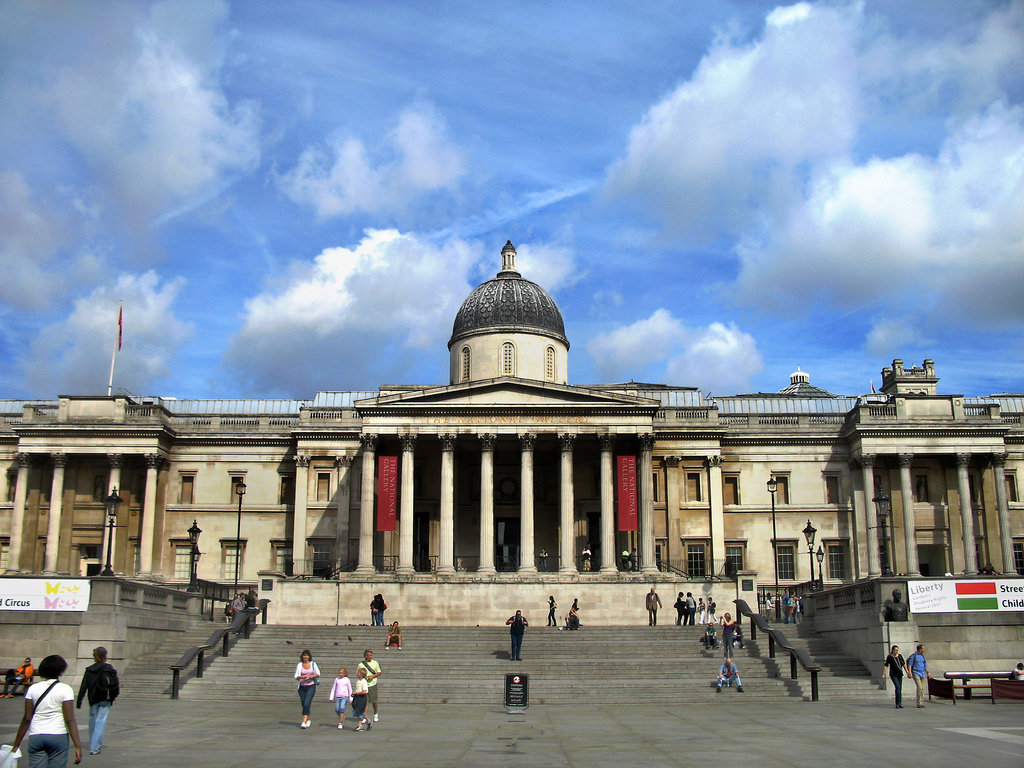} & & \figr[.135]{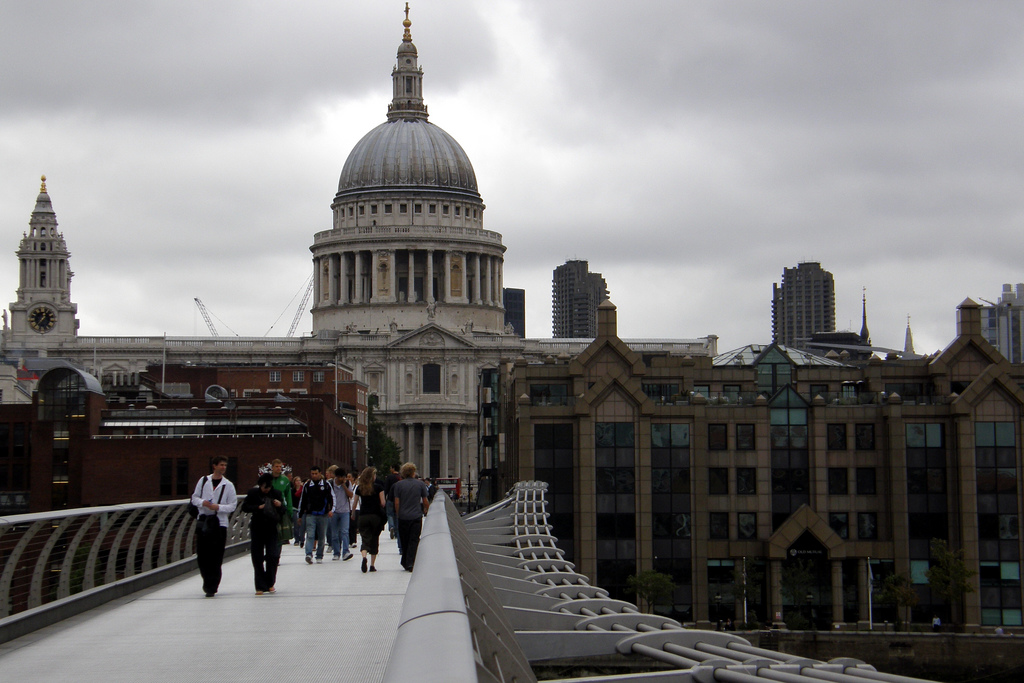} &\figr[.135]{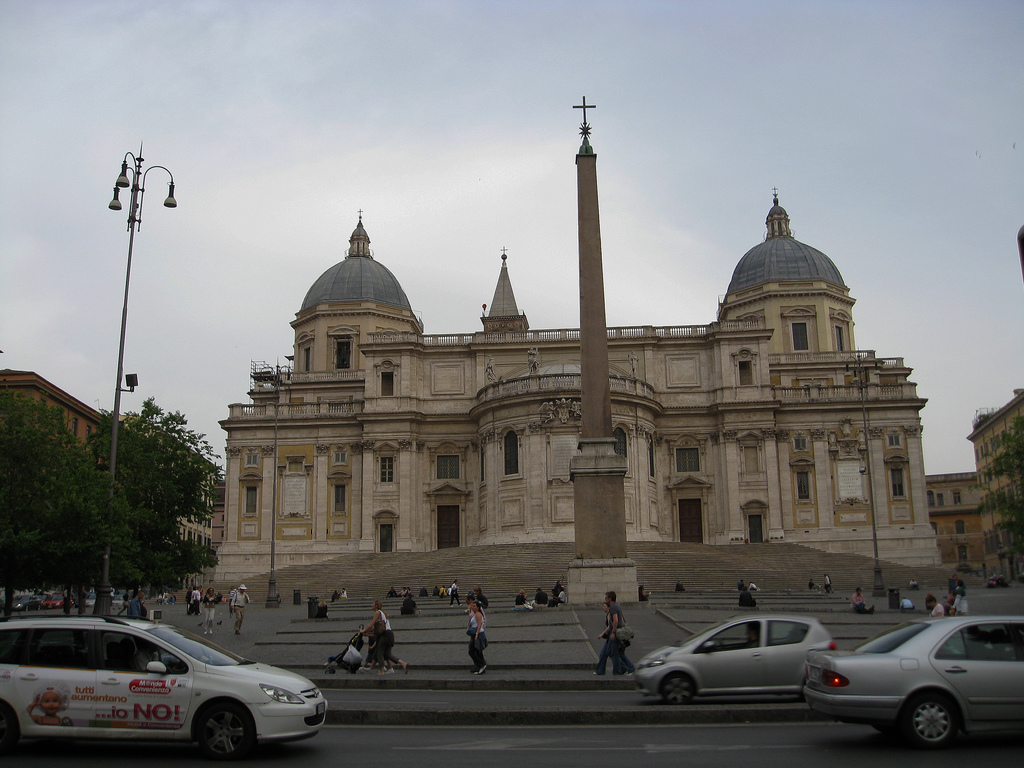} & \figr[.135]{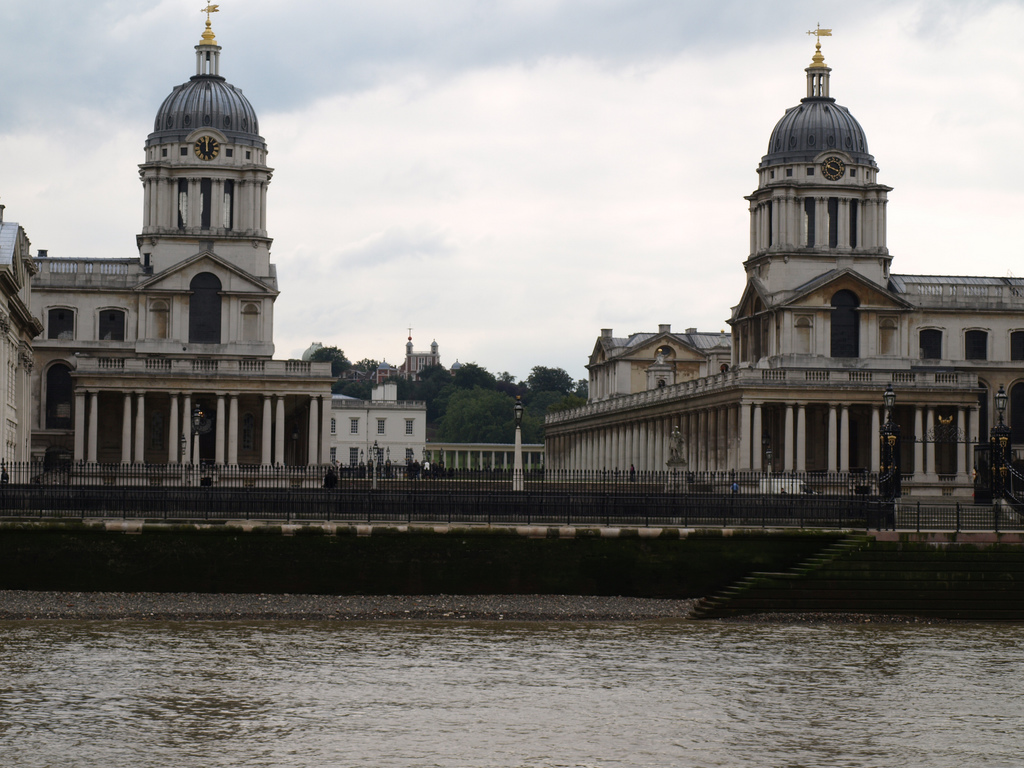} &\figr[.135]{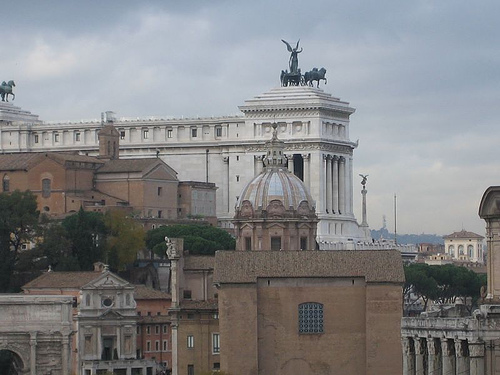} & \figr[.135]{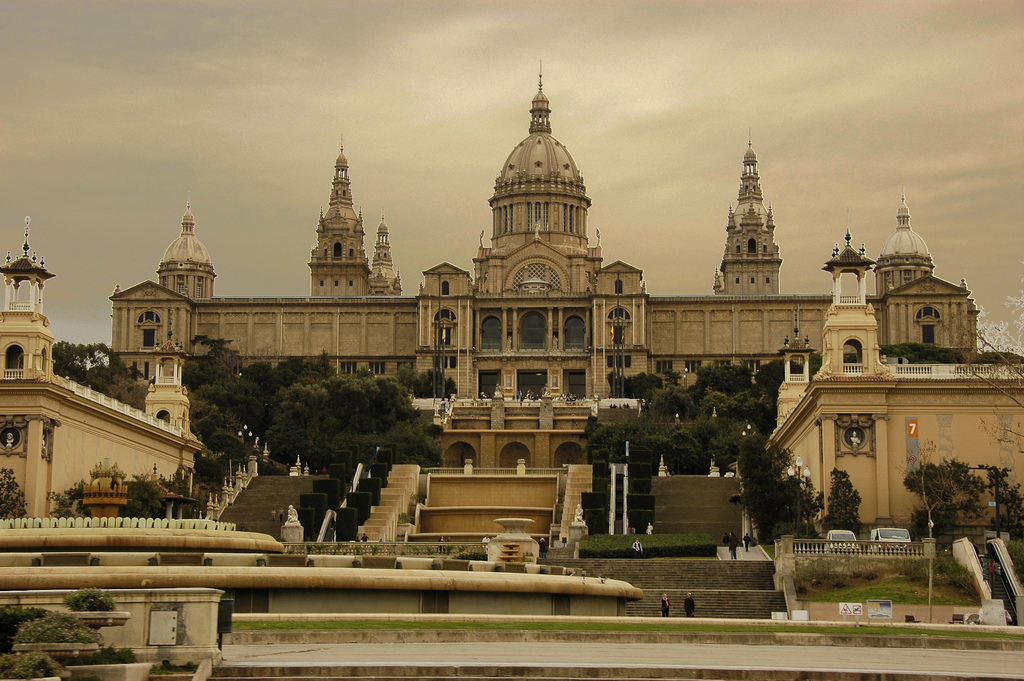} \\
\raisebox{3.5\height}{300} & \figr[.135]{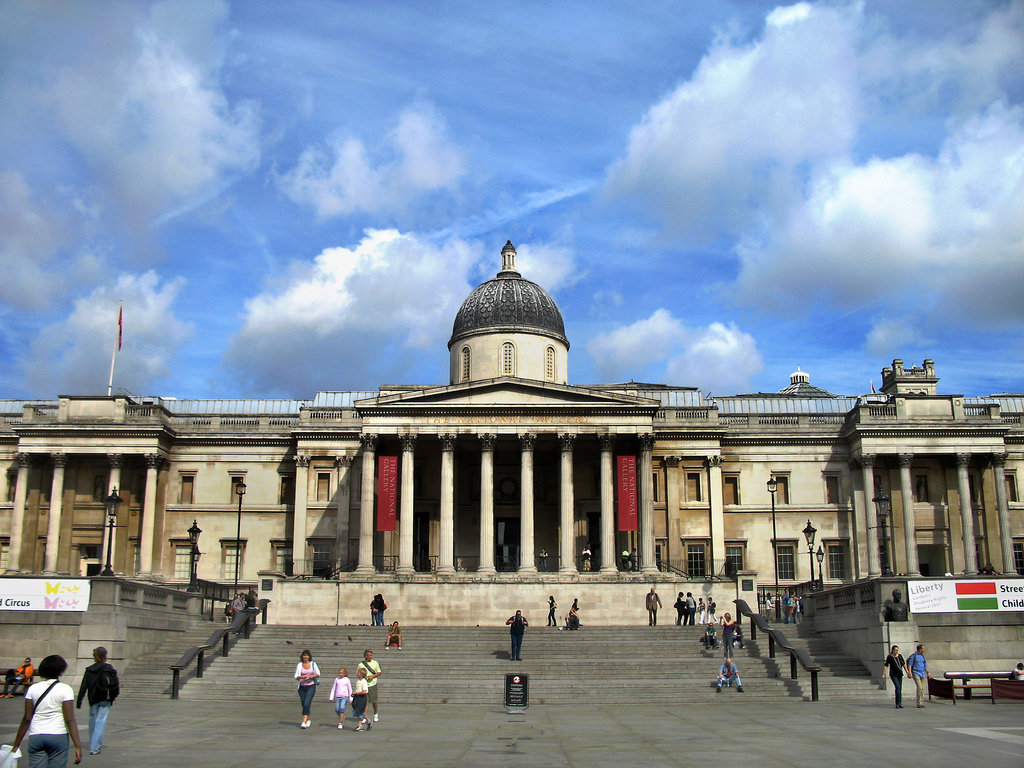} & & \figr[.135]{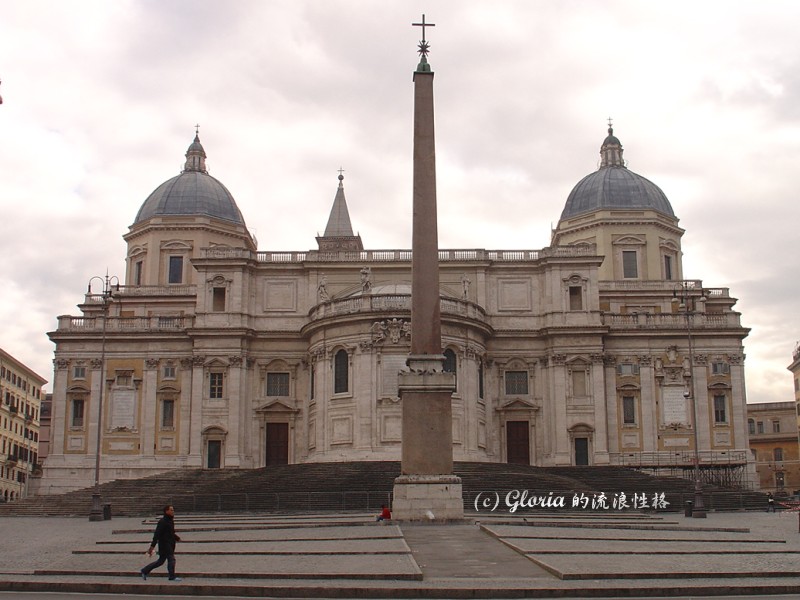} &\figr[.135]{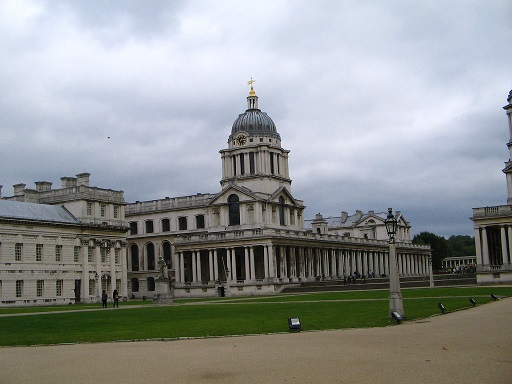} & \figr[.135]{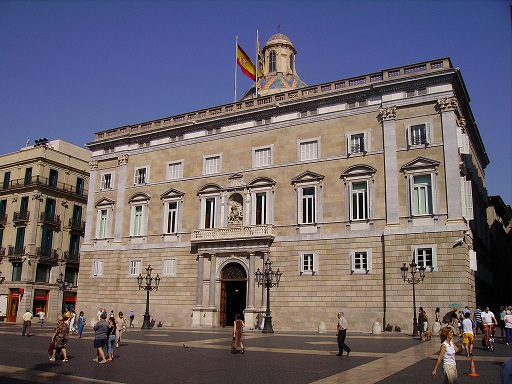} &\figr[.135]{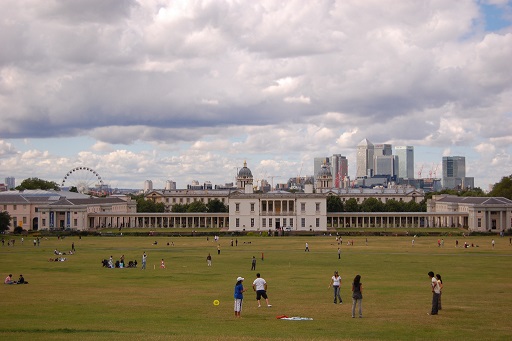} & \figr[.135]{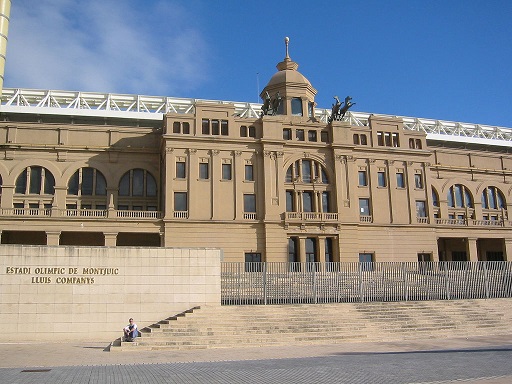} \\
\end{tabular}
\endgroup
\caption{\emph{Asymmetric hard negative mining}. One anchor $a$ shown on the first column, followed by the hard negatives $N(a)$ mined for this anchor, over different epochs. The anchor is represented by the student and the database of potential negatives by the teacher.}
\label{fig:asym_mining}
\end{figure}

Depending on the similarity we use in the loss function, \ie, symmetric~\eq{sym} or asymmetric\eq{asym}, we follow the same choice for hard negative mining. This means that, in mining based on \emph{asymmetric similarity}, the features of anchor $a$ come from the student model $f_\theta(a)$ while the database is represented by the teacher $g$. The database is fixed and does not need to be re-computed after each epoch. Only the anchors are updated, which makes training more efficient. \autoref{fig:asym_mining} gives an example of the hard negatives mined for one anchor across different epochs. Before the training starts (epoch 0) the results are not very informative, which is not surprising giving that the feature spaces of the teacher and the student do not match. However, after just a few epochs we see harder negatives being selected. This example illustrates how asymmetric similarity acts as a knowledge transfer mechanism from the teacher to the student model.

%------------------------------------------------------------------------------

\section{Model complexity and parameters}
\label{sec:comp_params}

\autoref{tab:net_params} gives the number of parameters and computational complexity (in FLOPS) for the networks used in this work. This includes two teachers. It is important to note that the versions of the teachers are already for image retrieval by removing the fully connected layers, hence this version of VGG16 has significantly fewer parameters than the original (around 138M). For the student networks, after removing the fully connected layers, a $1\times1$ convolutional layer is added to match the output dimensionality of the teacher network. Those models are shown in  \autoref{tab:net_params} as well as a version of the student model without the additional convolutional layer and fully connected layers.

%------------------------------------------------------------------------------
\begin{table}[h!]
\centering
\scriptsize %\footnotesize
\begin{tabular}{lcccc}
	\toprule
	\Th{Network} & \Th{Teacher} & \Th{d} & \Th{GFLOPS} & \Th{Param(M)}  \\ \midrule
	
	VGG16~\cite{radenovic2018fine}        &              & 512  & 79.40 & 14.71 \\
	ResNet101~\cite{radenovic2018fine}    &              & 2048 & 42.85 & 42.50 \\ \midrule
	\mr{3}{MobileNetV2}                   &              & 1280 &  1,74 &  2.22 \\
	                                      & VGG16        & 512  &  1.94 &  2.88 \\
	                                      & ResNet101    & 2048 &  2.50 &  4.85 \\ \cmidrule{2-5}
    \mr{3}{EfficientNet-B3}               &              & 1536 &  5.36 & 10.70  \\
                                          & VGG16        & 512  &  5.56 & 11.48 \\
	                                      & ResNet101    & 2048 &  6.26 & 13.84 \\
 	\bottomrule
\end{tabular}
\vspace{3pt}
\caption{FLOPS and parameters for the networks used in this work. In the top segment, the teacher networks are adapted for image retrieval, i.e. the fully connected layers are removed. The bottom segment shows the student network adapted in the same way and also with an added layer (or not) to match the output dimensionality of the teacher.}
\label{tab:net_params}
\end{table}
%------------------------------------------------------------------------------

\section{More results}
\label{sec:more}

\paragraph{Complete contrastive--regression ablation}

Here, we present the full version of the results of the ablation from \autoref{tab:sym_test_ablation}, for all four student-teacher combinations. Apart from mAP, we also report mP@10. \autoref{tab:sym_test_ablation_full} and \autoref{tab:asym_test_ablation_full} present the symmetric and asymmetric testing results, respectively. All results agree with the results of \autoref{tab:sym_test_ablation}. For symmetric testing, contrastive loss with a single positive and no negatives is again the worst. The addition of the anchor as a positive for itself as well as the negatives improve the results substantially. \lcontrp, which uses both, performs best in most cases with the exception of VGG16$\to$EfficientNet. For asymmetric testing, regression is the best. The inclusion of the anchor as positive for itself gives better results than without it.

%------------------------------------------------------------------------------

%------------------------------------------------------------------------------
\begin{table}
\centering
\scriptsize %\footnotesize
\setlength\tabcolsep{1.5pt}
\begin{tabular}{lcccccccccccccccccc}
	\toprule
	\mr{3}{\Th{Student}} && \mr{3}{$d$} & \mr{3}{\Th{Teacher}} & \mr{3}{\Th{Lab}} & \mr{3}{\Th{Loss}} & \mr{3}{\Th{Self}} & \mr{3}{\Th{Pos}} & \mr{3}{\Th{Neg}} & \mr{3}{\Th{Mining}} & \mc{4}{\Th{Medium}} & \mc{4}{\Th{Hard}} \\
	&&&&&&&&&& \mc{2}{\roxf} & \mc{2}{\rpar} & \mc{2}{\roxf} & \mc{2}{\rpar} \\
	&&&&&&&&&& {\scriptsize mAP} & {\scriptsize mP@10} & {\scriptsize mAP} & {\scriptsize mP@10} & {\scriptsize mAP} & {\scriptsize mP@10} & {\scriptsize mAP} & {\scriptsize mP@10} \\ \midrule
	\mr{12}{MobileNetV2}     && \mr{5}{512}  & \mr{5}{VGG16}     & \ch & \lcontr &     & \ch & \ch & hard & \tb{57.3} & 77.1 & 67.1      & 95.7 & 31.1      & 47.3 & 41.3      & 80.4 \\
	                         &&              &                   & \ch & \lcontr & \ch & \ch & \ch & hard & \tb{57.3} & 78.4 & \tb{68.4} & 96.1 & \tb{31.5} & 46.9 & \tb{42.2} & 78.9 \\
	                         &&              &                   & \ch & \lcontr &     & \ch &     & hard & 55.9      & 79.2 & 66.7      & 95.0 & 31.1      & 44.0 & 40.6      & 78.9 \\
	                         &&              &                   & \ch & \lcontr & \ch & \ch &     & hard & 55.5      & 76.1 & 67.0      & 96.0 & 30.4      & 44.1 & 40.9      & 81.4 \\ \cmidrule{5-18}
	                         &&              &                   &     & \lreg   & \ch &     &     & --   & 53.3      & 75.1 & 67.5      & 95.6 & 28.9      & 43.6 & 40.9      & 81.3 \\ \cmidrule{3-18}
	                         && \mr{5}{2048} & \mr{5}{ResNet101} & \ch & \lcontr &     & \ch & \ch & hard & 60.8      & 81.7 & 72.1      & 97.3 & 36.1      & 50.4 & 47.6      & 85.1 \\
	                         &&              &                   & \ch & \lcontr & \ch & \ch & \ch & hard & \tb{63.2} & 84.4 & \tb{75.0} & 98.0 & \tb{37.9} & 52.1 & \tb{52.0} & 87.3 \\
	                         &&              &                   & \ch & \lcontr &     & \ch &     & hard & 51.8      & 72.5 & 67.6      & 96.0 & 27.6      & 38.1 & 41.3      & 80.0 \\
	                         &&              &                   & \ch & \lcontr & \ch & \ch &     & hard & 60.6      & 80.0 & 74.1      & 97.0 & 35.7      & 49.4 & 50.9      & 85.6 \\ \cmidrule{5-18}
	                         &&              &                   &     & \lreg   & \ch &     &     & --   & 59.8      & 80.3 & 73.1      & 96.9 & 35.7      & 49.4 & 49.5      & 84.7 \\ \midrule
	\mr{12}{EfficientNet-B3} && \mr{5}{512}  & \mr{5}{VGG16}     & \ch & \lcontr &     & \ch & \ch & hard & 56.8      & 75.7 & \tb{70.4} & 96.3 & 31.2      & 43.9 & \tb{45.4} & 81.7 \\
	                         &&              &                   & \ch & \lcontr & \ch & \ch & \ch & hard & 56.9      & 75.6 & 69.0      & 96.0 & 31.1      & 46.7 & 43.5      & 80.9 \\
	                         &&              &                   & \ch & \lcontr &     & \ch &     & hard & 56.1      & 77.0 & 69.3      & 96.4 & 30.1      & 42.1 & 44.7      & 78.4 \\
	                         &&              &                   & \ch & \lcontr & \ch & \ch &     & hard & \tb{57.6} & 78.3 & 69.9      & 96.9 & \tb{31.4} & 46.7 & 44.9      & 82.6 \\ \cmidrule{5-18}
	                         &&              &                   &     & \lreg   & \ch &     &     & --   & 55.0      & 75.0 & 69.4      & 96.6 & 27.1      & 42.3 & 44.5      & 80.4 \\ \cmidrule{3-18}
	                         && \mr{5}{2048} & \mr{5}{ResNet101} & \ch & \lcontr &     & \ch & \ch & hard & 66.3      & 85.3 & \tb{77.4} & 98.4 & 41.3      & 58.9 & \tb{55.5} & 88.3 \\
	                         &&              &                   & \ch & \lcontr & \ch & \ch & \ch & hard & \tb{66.8} & 84.7 & 77.1      & 98.6 & \tb{42.5} & 58.7 & \tb{55.5} & 87.9 \\
	                         &&              &                   & \ch & \lcontr &     & \ch &     & hard & 61.7      & 81.7 & 74.3      & 97.1 & 36.1      & 51.7 & 51.6      & 85.9 \\
	                         &&              &                   & \ch & \lcontr & \ch & \ch &     & hard & 63.8      & 83.1 & 75.9      & 98.3 & 40.1      & 54.3 & 54.4      & 87.1 \\ \cmidrule{5-18}
	                         &&              &                   &     & \lreg   & \ch &     &     & --   & 64.9      & 83.7 & 74.4      & 97.7 & 40.5      & 55.9 & 52.4      & 87.1 \\
	\bottomrule
\end{tabular}
\vspace{3pt}
\caption{\emph{Complete contrastive--regression ablation: symmetric testing} on \roxf and \rpar~\cite{RIT+18}. \Th{Lab}: using labels in student model training. \Th{Pos}, \Th{Neg}: Using positives, negatives. \Th{Self}: Using anchor (by teacher) as positive for itself (by student). Using asymmetric similarity~\eq{asym} at training in all cases. Best mAP highlighted per teacher-student pair. GeM pooling and learned whitening~\cite{radenovic2018fine} used in all cases.}
\label{tab:sym_test_ablation_full}
\end{table}
%------------------------------------------------------------------------------

%------------------------------------------------------------------------------
\begin{table}
\centering
\scriptsize %\footnotesize
\setlength\tabcolsep{1.5pt}
\begin{tabular}{lccccccccccccccccccc}
	\toprule
	\mr{3}{\Th{Student}} && \mr{3}{$d$} & \mr{3}{\Th{Teacher}} & \mr{3}{\Th{Lab}} & \mr{3}{\Th{Loss}} & \mr{3}{\Th{Self}} & \mr{3}{\Th{Pos}} & \mr{3}{\Th{Neg}} & \mr{3}{\Th{Mining}} & \mc{4}{\Th{Medium}} & \mc{4}{\Th{Hard}} \\
	&&&&&&&&&& \mc{2}{\roxf} & \mc{2}{\rpar} & \mc{2}{\roxf} & \mc{2}{\rpar} \\
	&&&&&&&&&& {\scriptsize mAP} & {\scriptsize mP@10} & {\scriptsize mAP} & {\scriptsize mP@10} & {\scriptsize mAP} & {\scriptsize mP@10} & {\scriptsize mAP} & {\scriptsize mP@10} \\ \midrule
	\mr{12}{MobileNetV2}     && \mr{5}{512}  & \mr{5}{VGG16}     & \ch & \lcontr &     & \ch & \ch & hard & 38.3      & 53.7 & 49.8      & 84.4 & 18.4      & 32.8 & 23.8      & 55.7 \\
	                         &&              &                   & \ch & \lcontr & \ch & \ch & \ch & hard & 42.9      & 59.1 & 55.9      & 88.4 & 22.6      & 35.2 & 31.4      & 66.3 \\
	                         &&              &                   & \ch & \lcontr &     & \ch &     & hard & 34.1      & 48.9 & 47.3      & 82.0 & 17.0      & 25.6 & 24.5      & 53.4 \\
	                         &&              &                   & \ch & \lcontr & \ch & \ch &     & hard & 38.2      & 52.0 & 52.2      & 86.0 & 15.3      & 26.2 & 28.9      & 64.1 \\ \cmidrule{5-18}
	                         &&              &                   &     & \lreg   & \ch &     &     & --   & \tb{48.0} & 64.3 & \tb{57.9} & 90.7 & \tb{26.5} & 37.9 & \tb{32.6} & 67.1 \\ \cmidrule{3-18}
	                         && \mr{5}{2048} & \mr{5}{ResNet101} & \ch & \lcontr &     & \ch & \ch & hard & 32.3      & 49.7 & 51.5      & 83.3 &  9.6      & 18.3 & 28.2      & 62.4  \\
	                         &&              &                   & \ch & \lcontr & \ch & \ch & \ch & hard & 47.1      & 65.4 & 61.5      & 92.6 & 21.8      & 33.1 & 37.7      & 74.1 \\
	                         &&              &                   & \ch & \lcontr &     & \ch &     & hard & 27.3      & 38.4 & 47.7      & 80.9 &  8.4      & 15.3 & 24.3      & 50.6 \\
	                         &&              &                   & \ch & \lcontr & \ch & \ch &     & hard & 40.5      & 58.2 & 55.8      & 87.6 & 17.4      & 26.3 & 29.9      & 63.4 \\ \cmidrule{5-18}
	                         &&              &                   &     & \lreg   & \ch &     &     & --   & \tb{49.2} & 67.9 & \tb{65.0} & 92.6 & \tb{23.3} & 36.9 & \tb{40.7} & 72.1 \\ \midrule
	\mr{12}{EfficientNet-B3} && \mr{5}{512}  & \mr{5}{VGG16}     & \ch & \lcontr &     & \ch & \ch & hard & 43.8      & 74.7 & 24.9      & 39.3 & 23.0      & 51.3 &  6.1      & 15.6 \\
	                         &&              &                   & \ch & \lcontr & \ch & \ch & \ch & hard & 44.7      & 61.5 & 58.0      & 93.3 & 23.9      & 37.9 & 32.4      & 69.1 \\
	                         &&              &                   & \ch & \lcontr &     & \ch &     & hard & 32.4      & 45.4 & 47.8      & 84.4 & 14.1      & 22.0 & 25.8      & 56.3 \\
	                         &&              &                   & \ch & \lcontr & \ch & \ch &     & hard & 41.6      & 57.5 & 53.9      & 90.1 & 20.3      & 30.6 & 30.2      & 64.0 \\ \cmidrule{5-18}
	                         &&              &                   &     & \lreg   & \ch &     &     & --   & \tb{49.4} & 70.0 & \tb{58.2} & 92.4 & \tb{26.0} & 39.6 & \tb{33.0} & 70.6 \\ \cmidrule{3-18}
	                         && \mr{5}{2048} & \mr{5}{ResNet101} & \ch & \lcontr &     & \ch & \ch & hard & 37.4      & 56.8 & 57.4      & 90.4 & 10.9      & 24.6 & 33.7      & 65.9 \\
	                         &&              &                   & \ch & \lcontr & \ch & \ch & \ch & hard & 45.2      & 67.2 & 63.7      & 92.1 & 19.6      & 35.5 & 40.9      & 73.6 \\
	                         &&              &                   & \ch & \lcontr &     & \ch &     & hard & 30.8      & 44.5 & 51.2      & 83.7 & 10.2      & 16.1 & 27.8      & 57.0 \\
	                         &&              &                   & \ch & \lcontr & \ch & \ch &     & hard & 40.1      & 56.7 & 59.1      & 91.1 & 14.6      & 24.3 & 35.0      & 71.0 \\ \cmidrule{5-18}
	                         &&              &                   &     & \lreg   & \ch &     &     & --   & \tb{52.9} & 71.8 & \tb{65.2} & 93.3 & \tb{27.8} & 41.5 & \tb{42.4} & 71.9 \\
	\bottomrule
\end{tabular}
\vspace{3pt}
\caption{\emph{Complete contrastive--regression ablation: asymmetric testing} on \roxf and \rpar~\cite{RIT+18}. \Th{Lab}: using labels in student model training. \Th{Pos}, \Th{Neg}: Using positives, negatives. \Th{Self}: Using anchor (by teacher) as positive for itself (by student). Using asymmetric similarity~\eq{asym} at training in all cases. Best mAP highlighted per teacher-student pair. GeM pooling and learned whitening~\cite{radenovic2018fine} used in all cases.}
\label{tab:asym_test_ablation_full}
\end{table}
%------------------------------------------------------------------------------

%------------------------------------------------------------------------------
\begin{table}
\centering
\scriptsize %\footnotesize
\setlength\tabcolsep{1.5pt}
\begin{tabular}{lcccccccccccccccc}
	\toprule
	\mr{3}{\Th{Student}} && \mr{3}{$d$} & \mr{3}{\Th{Teacher}} && \mr{3}{\Th{Lab}} & \mr{3}{\Th{Loss}} & \mr{3}{\Th{Mining}} && \mc{4}{\Th{Medium}} & \mc{4}{\Th{Hard}} \\
	&&&&&&&& \Th{Asym} & \mc{2}{\ro} & \mc{2}{\rp} & \mc{2}{\ro} & \mc{2}{\rp} \\
	&&&&&&&&& {\scriptsize mAP} & {\scriptsize mP@10} & {\scriptsize mAP} & {\scriptsize mP@10} & {\scriptsize mAP} & {\scriptsize mP@10} & {\scriptsize mAP} & {\scriptsize mP@10} \\ \midrule
	VGG16                    && 512          &                   && \ch & \lcontr  & hard   &     & 60.9      & 81.9 & 69.3      & 97.4 & 32.9      & 50.9 & 44.2      & 83.1 \\
	ResNet101                && 2048         &                   && \ch & \lcontr  & hard   &     & 65.4      & 85.7 & 76.7      & 98.4 & 40.1      & 56.6 & 55.2      & 87.7 \\
	\mr{2}{MobileNetV2}      && 512          &                   && \ch & \lcontr  & hard   &     & 53.6      & 75.8 & 66.4      & 96.7 & 28.8      & 42.9 & 39.7      & 79.0 \\
	                         && 2048         &                   && \ch & \lcontr  & hard   &     & 56.1      & 79.0 & 68.5      & 98.1 & 30.3      & 46.0 & 42.0      & 82.6 \\
	\mr{2}{EfficientNet-B3}  && 512          &                   && \ch & \lcontr  & hard   &     & 53.8      & 76.6 & 70.9      & 96.6 & 26.2      & 42.3 & 46.0      & 83.7 \\
	                         && 2048         &                   && \ch & \lcontr  & hard   &     & 59.6      & 86.1 & 75.1      & 95.1 & 33.3      & 46.0 & 51.9      & 87.6 \\ \midrule
	\mr{16}{MobileNetV2}     && \mr{8}{512}  & \mr{8}{VGG16}     && \ch & \lcontrp & hard   & \ch & \tb{57.3} & 78.4 & \tb{68.4} & 96.1 & \tb{31.5} & 46.9 & \tb{42.2} & 78.9 \\
	                         &&              &                   && \ch & \lcontr  & hard   & \ch & \tb{57.3} & 77.1 & 67.1      & 95.7 & 31.1      & 47.3 & 41.3      & 80.4 \\
	                         &&              &                   && \ch & \ltripl  & hard   & \ch & 37.0      & 55.2 & 62.4      & 94.4 & 11.8      & 23.0 & 36.0      & 73.7 \\
	                         &&              &                   && \ch & \lms     & hard   & \ch & 36.8      & 55.2 & 62.8      & 94.4 & 11.5      & 22.2 & 36.5      & 75.0 \\ \cmidrule{5-17}
	                         &&              &                   &&     & \lreg    & --     & \ch & 53.3      & 75.1 & 67.5      & 95.6 & 28.9      & 43.6 & 40.9      & 81.3 \\
	                         &&              &                   &&     & \lrkd    & random &     & 46.2      & 68.1 & 64.3      & 94.7 & 21.8      & 32.8 & 37.6      & 72.3 \\
	                         &&              &                   &&     & \lrank   & random &     & 45.2      & 66.5 & 60.6      & 92.1 & 24.6      & 34.9 & 33.1      & 74.1 \\ \cmidrule{3-17}
	                         && \mr{8}{2048} & \mr{8}{ResNet101} && \ch & \lcontrp & hard   & \ch & \tb{63.2} & 84.4 & \tb{75.0} & 98.0 & \tb{37.9} & 52.1 & \tb{52.0} & 87.3 \\
	                         &&              &                   && \ch & \lcontr  & hard   & \ch & 60.8      & 81.7 & 72.1      & 97.3 & 36.1      & 50.4 & 47.6      & 85.1 \\
	                         &&              &                   && \ch & \ltripl  & hard   & \ch & 45.5      & 66.1 & 68.0      & 96.1 & 19.6      & 33.5 & 43.4      & 80.6 \\
	                         &&              &                   && \ch & \lms     & hard   & \ch & 44.5      & 65.4 & 68.1      & 96.1 & 17.9      & 32.1 & 43.2      & 80.1 \\ \cmidrule{5-17}
	                         &&              &                   &&     & \lreg    & --     & \ch & 59.8      & 80.3 & 73.1      & 96.9 & 35.7      & 49.4 & 49.5      & 84.7 \\
	                         &&              &                   &&     & \lrkd    & random &     & 56.1      & 79.3 & 69.8      & 96.3 & 31.8      & 46.0 & 44.2      & 82.3 \\
	                         &&              &                   &&     & \lrank   & random &     & 48.3      & 69.8 & 58.0      & 94.3 & 23.3      & 33.3 & 31.5      & 71.4 \\ \midrule
	\mr{16}{EfficientNet-B3} && \mr{8}{512}  & \mr{8}{VGG16}     && \ch & \lcontrp & hard   & \ch & \tb{56.9} & 75.6 & 69.0      & 96.0 & 31.1      & 46.7 & 43.5      & 80.9 \\
	                         &&              &                   && \ch & \lcontr  & hard   & \ch & 56.8      & 75.7 & \tb{70.4} & 96.3 & \tb{31.2} & 43.9 & \tb{45.4} & 81.7 \\
	                         &&              &                   && \ch & \ltripl  & hard   & \ch & 33.7      & 48.5 & 64.6      & 94.4 & 8.0       & 20.1 & 40.3      & 76.1 \\
	                         &&              &                   && \ch & \lms     & hard   & \ch & 33.9      & 49.5 & 64.9      & 94.4 &  8.1      & 20.4 & 40.6      & 76.9 \\ \cmidrule{5-17}
	                         &&              &                   &&     & \lreg    & --     & \ch & 55.0      & 75.0 & 69.4      & 96.6 & 27.1      & 42.3 & 44.5      & 80.4 \\
	                         &&              &                   &&     & \lrkd    & random &     & 51.6      & 71.4 & 67.6      & 95.3 & 26.2      & 38.5 & 41.7      & 81.1 \\
	                         &&              &                   &&     & \lrank   & random &     & 34.7      & 50.0 & 61.1      & 93.1 &  8.5      & 17.8 & 35.2      & 67.1 \\ \cmidrule{3-17}
	                         && \mr{8}{2048} & \mr{8}{ResNet101} && \ch & \lcontrp & hard   & \ch & \tb{66.8} & 84.7 & 77.1      & 98.6 & \tb{42.5} & 58.7 & \tb{55.5} & 87.9 \\
	                         &&              &                   && \ch & \lcontr  & hard   & \ch & 66.3      & 85.3 & \tb{77.4} & 98.4 & 41.3      & 58.9 & \tb{55.5} & 88.3 \\
	                         &&              &                   && \ch & \ltripl  & hard   & \ch & 39.5      & 57.3 & 69.4      & 95.9 & 11.6      & 24.3 & 45.8      & 81.1 \\
	                         &&              &                   && \ch & \lms     & hard   & \ch & 39.9      & 57.4 & 69.7      & 95.7 & 11.7      & 24.2 & 46.2      & 81.4  \\ \cmidrule{5-17}
	                         &&              &                   &&     & \lreg    & --     & \ch & 64.9      & 83.7 & 74.4      & 97.7 & 40.5      & 55.9 & 52.4      & 87.1 \\
	                         &&              &                   &&     & \lrkd    & random &     & 56.3      & 75.8 & 73.0      & 98.4 & 30.5      & 46.4 & 46.4      & 82.3 \\
	                         &&              &                   &&     & \lrank   & random &     & 40.3      & 58.4 & 69.9      & 95.9 & 11.8      & 24.6 & 46.4      & 81.1 \\
	\bottomrule
\end{tabular}
\vspace{3pt}
\caption{\emph{Symmetric testing} on \roxf and \rpar~\cite{RIT+18}. \Th{Lab}: using labels in student model training. \Th{Asym}: Using asymmetric similarity~\eq{asym} at training. Best mAP highlighted per teacher-student pair. GeM pooling and learned whitening~\cite{radenovic2018fine} used in all cases.}
\label{tab:sym_test_full}
\end{table}
%------------------------------------------------------------------------------

\paragraph{Complete symmetric and asymmetric testing results}

\autoref{tab:sym_test_full} supplements \autoref{tab:sym_asym_test_full} by adding mP@10 scores for all the symmetric testing experiments. Similarly, \autoref{tab:asym_test_full} adds mP@10 results to all asymmetric testing experiments. Overall, the conclusions made based on mAP apply to the mP@10 results.

%------------------------------------------------------------------------------
\begin{table}
\centering
\scriptsize %\footnotesize
\setlength\tabcolsep{1.5pt}
\begin{tabular}{lcccccccccccccccc}
	\toprule
	\mr{3}{\Th{Student}} && \mr{3}{$d$} & \mr{3}{\Th{Teacher}} && \mr{3}{\Th{Lab}} & \mr{3}{\Th{Loss}} & \mr{3}{\Th{Mining}} && \mc{4}{\Th{Medium}} & \mc{4}{\Th{Hard}} \\
	&&&&&&&& \Th{Asym} &  \mc{2}{\ro} & \mc{2}{\rp} & \mc{2}{\ro} & \mc{2}{\rp} \\
	&&&&&&&&& {\scriptsize mAP} & {\scriptsize mP@10} & {\scriptsize mAP} & {\scriptsize mP@10} & {\scriptsize mAP} & {\scriptsize mP@10} & {\scriptsize mAP} & {\scriptsize mP@10} \\ \midrule
	VGG16                    && 512          &                   && \ch & \lcontr  & hard   &     & 60.9      & 81.9 & 69.3      & 97.4 & 32.9      & 50.9 & 44.2      & 83.1 \\
	ResNet101                && 2048         &                   && \ch & \lcontr  & hard   &     & 65.4      & 85.7 & 76.7      & 98.4 & 40.1      & 56.6 & 55.2      & 87.7 \\
	\mr{2}{MobileNetV2}      && 512          &                   && \ch & \lcontr  & hard   &     & 53.6      & 75.8 & 66.4      & 96.7 & 28.8      & 42.9 & 39.7      & 79.0 \\
	                         && 2048         &                   && \ch & \lcontr  & hard   &     & 56.1      & 79.0 & 68.5      & 98.1 & 30.3      & 46.0 & 42.0      & 82.6 \\
	\mr{2}{EfficientNet-B3}  && 512          &                   && \ch & \lcontr  & hard   &     & 53.8      & 76.6 & 70.9      & 96.6 & 26.2      & 42.3 & 46.0      & 83.7 \\
	                         && 2048         &                   && \ch & \lcontr  & hard   &     & 59.6      & 86.1 & 75.1      & 95.1 & 33.3      & 46.0 & 51.9      & 87.6 \\ \midrule
	\mr{16}{MobileNetV2}     && \mr{8}{512}  & \mr{8}{VGG16}     && \ch & \lcontrp & hard   & \ch & 42.9      & 59.1 & 55.9      & 88.4 & 22.6      & 35.2 & 31.4      & 66.3 \\
	                         &&              &                   && \ch & \lcontr  & hard   & \ch & 38.3      & 53.7 & 49.8      & 84.4 & 18.4      & 32.8 & 23.8      & 55.7 \\
	                         &&              &                   && \ch & \ltripl  & hard   & \ch &  1.8      &  0.0 &  4.3      &  1.3 &  0.7      &  0.0 &  2.8      &  1.4 \\
	                         &&              &                   && \ch & \lms     & hard   & \ch &  1.9      &  0.0 &  4.3      &  1.6 &  0.8      &  0.0 &  2.7      &  1.6 \\ \cmidrule{5-17}
	                         &&              &                   &&     & \lreg    & --     & \ch & \tb{48.0} & 64.3 & \tb{57.9} & 90.7 & \tb{26.5} & 37.9 & \tb{32.6} & 67.1 \\
	                         &&              &                   &&     & \lrkd    & random &     &  2.0      &  0.0 &  4.1      &  1.0 &  0.8      &  0.0 &  2.6      &  1.0 \\
	                         &&              &                   &&     & \lrank   & random &     &  1.7      &  0.0 &  3.8      &  0.3 &  0.7      &  0.0 &  2.4      &  0.3 \\ \cmidrule{3-17}
	                         && \mr{8}{2048} & \mr{8}{ResNet101} && \ch & \lcontrp & hard   & \ch & 47.1      & 65.4 & 61.5      & 92.6 & 21.8      & 33.1 & 37.7      & 74.1 \\
	                         &&              &                   && \ch & \lcontr  & hard   & \ch & 32.3      & 49.7 & 51.5      & 83.3 &  9.6      & 18.3 & 28.2      & 62.4 \\
	                         &&              &                   && \ch & \ltripl  & hard   & \ch &  1.3      &  0.0 &  3.7      &  1.4 &  0.7      &  0.0 &  2.4      &  1.4 \\
	                         &&              &                   && \ch & \lms     & hard   & \ch &  1.4      &  0.3 &  3.6      &  1.0 &  0.7      &  0.3 &  2.3      &  0.9 \\ \cmidrule{5-17}
	                         &&              &                   &&     & \lreg    & --     & \ch & \tb{49.2} & 67.9 & \tb{65.0} & 92.6 & \tb{23.3} & 36.9 & \tb{40.7} & 72.1 \\
	                         &&              &                   &&     & \lrkd    & random &     &  1.6      &  1.3 &  4.1      &  2.3 &  0.8      &  1.1 &  2.5      &  1.6 \\
	                         &&              &                   &&     & \lrank   & random &     &  1.3      &  0.4 &  4.1      &  3.6 &  0.6      &  0.3 &  2.7      &  3.1 \\ \midrule
	\mr{16}{EfficientNet-B3} && \mr{8}{512}  & \mr{8}{VGG16}     && \ch & \lcontrp & hard   & \ch & 44.7      & 61.5 & 58.0      & 93.3 & 23.9      & 37.9 & 32.4      & 69.1 \\
	                         &&              &                   && \ch & \lcontr  & hard   & \ch & 43.8      & 74.7 & 24.9      & 39.3 & 23.0      & 51.3 &  6.1      & 15.6 \\
	                         &&              &                   && \ch & \ltripl  & hard   & \ch &  1.4      &  0.0 &  4.0      &  0.0 &  0.6      &  0.0 &  2.5      &  0.0 \\
	                         &&              &                   && \ch & \lms     & hard   & \ch &  1.4      &  0.0 &  3.9      &  0.0 &  0.6      &  0.0 &  2.5      &  0.0 \\ \cmidrule{5-17}
	                         &&              &                   &&     & \lreg    & --     & \ch & \tb{49.4} & 70.0 & \tb{58.2} & 92.4 & \tb{26.0} & 39.6 & \tb{33.0} & 70.6 \\
	                         &&              &                   &&     & \lrkd    & random &     &  1.3      &  0.0 &  3.8      &  0.7 &  0.6      &  0.0 &  2.5      &  0.3 \\
	                         &&              &                   &&     & \lrank   & random &     &  1.4      &  0.0 &  3.9      &  0.0 &  0.6      &  0.0 &  2.4      &  0.0 \\ \cmidrule{3-17}
	                         && \mr{8}{2048} & \mr{8}{ResNet101} && \ch & \lcontrp & hard   & \ch & 45.2      & 67.2 & 63.7      & 92.1 & 19.6      & 35.5 & 40.9      & 73.6 \\
	                         &&              &                   && \ch & \lcontr  & hard   & \ch & 37.4      & 56.8 & 57.4      & 90.4 & 10.9      & 24.6 & 33.7      & 65.9 \\
	                         &&              &                   && \ch & \ltripl  & hard   & \ch &  1.5      &  0.7 &  4.0      &  1.6 &  0.7      &  0.7 &  2.5      &  0.9 \\
	                         &&              &                   && \ch & \lms     & hard   & \ch &  1.5      &  0.7 &  4.0      &  1.4 &  0.7      &  0.7 &  2.4      &  1.0 \\ \cmidrule{5-17}
	                         &&              &                   &&     & \lreg    & --     & \ch & \tb{52.9} & 71.8 & \tb{65.2} & 93.3 & \tb{27.8} & 41.5 & \tb{42.4} & 71.9 \\
	                         &&              &                   &&     & \lrkd    & random &     &  1.6      &  0.7 &  3.8      &  1.6 &  0.7      &  0.4 &  2.4      &  0.7 \\
	                         &&              &                   &&     & \lrank   & random &     &  1.5      &  0.9 &  4.0      &  1.9 &  0.7      &  0.9 &  2.5      &  1.4 \\
	\bottomrule
\end{tabular}
\vspace{3pt}
\caption{\emph{Asymmetric testing} on \roxf and \rpar~\cite{RIT+18}. \Th{Lab}: using labels in student model training. \Th{Asym}: Using asymmetric similarity~\eq{asym} at training. Best mAP highlighted per teacher-student pair. GeM pooling and learned whitening~\cite{radenovic2018fine} used in all cases. The results without a teacher in the top block correspond to symmetric testing (same as in \autoref{tab:sym_test_full}) and are only added here for convenience.}
\label{tab:asym_test_full}
\end{table}
%------------------------------------------------------------------------------

%% file: main.bbl
\begin{thebibliography}{10}\itemsep=-1pt

\bibitem{ba2014deep}
Jimmy Ba and Rich Caruana.
\newblock Do deep nets really need to be deep?
\newblock In {\em NIPS}, 2014.

\bibitem{BSCL14}
Artem Babenko, Anton Slesarev, Alexandr Chigorin, and Victor Lempitsky.
\newblock Neural codes for image retrieval.
\newblock In {\em ECCV}, 2014.

\bibitem{BN03}
Mikhail Belkin and Partha Niyogi.
\newblock Laplacian eigenmaps for dimensionality reduction and data
  representation.
\newblock {\em Neural computation}, 15(6), 2003.

\bibitem{BoSI08}
Oren Boiman, Eli Shechtman, and Michal Irani.
\newblock In defense of nearest-neighbor based image classification.
\newblock In {\em CVPR}, 2008.

\bibitem{Cakir_2019_CVPR}
Fatih Cakir, Kun He, Xide Xia, Brian Kulis, and Stan Sclaroff.
\newblock Deep metric learning to rank.
\newblock In {\em CVPR}, 2019.

\bibitem{cao2019unsupervised}
Xuefei Cao, Bor-Chun Chen, and Ser-Nam Lim.
\newblock Unsupervised deep metric learning via auxiliary rotation loss.
\newblock {\em arXiv preprint arXiv:1911.07072}, 2019.

\bibitem{CQL+07}
Zhe Cao, Tao Qin, Tie-Yan Liu, Ming-Feng Tsai, and Hang Li.
\newblock Learning to rank: From pairwise approach to listwise approach.
\newblock In {\em ICML}, 2007.

\bibitem{chen2020simple}
Ting Chen, Simon Kornblith, Mohammad Norouzi, and Geoffrey Hinton.
\newblock A simple framework for contrastive learning of visual
  representations.
\newblock {\em arXiv preprint arXiv:2002.05709}, 2020.

\bibitem{chen2018darkrank}
Yuntao Chen, Naiyan Wang, and Zhaoxiang Zhang.
\newblock {DarkRank}: Accelerating deep metric learning via cross sample
  similarities transfer.
\newblock In {\em AAAI}, 2018.

\bibitem{DaBS14}
Damek Davis, Jonathan Balzer, and Stefano Soatto.
\newblock Asymmetric sparse kernel approximations for large-scale visual
  search.
\newblock In {\em CVPR}, 2014.

\bibitem{DGXZ19}
Jiankang Deng, Jia Guo, Niannan Xue, and Stefanos Zafeiriou.
\newblock {ArcFace}: Additive angular margin loss for deep face recognition.
\newblock In {\em CVPR}, 2019.

\bibitem{DoCL08}
Wei Dong, Moses Charikar, and Kai Li.
\newblock Asymmetric distance estimation with sketches for similarity search in
  high-dimensional spaces.
\newblock In {\em SIGIR}, 2008.

\bibitem{faghri2017vse++}
Fartash Faghri, David~J Fleet, Jamie~Ryan Kiros, and Sanja Fidler.
\newblock {VSE++}: Improving visual-semantic embeddings with hard negatives.
\newblock {\em arXiv preprint arXiv:1707.05612}, 2017.

\bibitem{frome2013devise}
Andrea Frome, Greg~S Corrado, Jon Shlens, Samy Bengio, Jeff Dean, Marc'Aurelio
  Ranzato, and Tomas Mikolov.
\newblock Devise: A deep visual-semantic embedding model.
\newblock In {\em NIPS}, 2013.

\bibitem{gong2014compressing}
Yunchao Gong, Liu Liu, Ming Yang, and Lubomir Bourdev.
\newblock Compressing deep convolutional networks using vector quantization.
\newblock {\em arXiv preprint arXiv:1412.6115}, 2014.

\bibitem{GARL16}
Albert Gordo, Jon Almazan, Jerome Revaud, and Diane Larlus.
\newblock Deep image retrieval: Learning global representations for image
  search.
\newblock {\em ECCV}, 2016.

\bibitem{GoPe11}
Albert Gordo and Florent Perronnin.
\newblock Asymmetric distances for binary embeddings.
\newblock In {\em CVPR}, 2011.

\bibitem{HaCL06}
Raia Hadsell, Sumit Chopra, and Yann Lecun.
\newblock Dimensionality reduction by learning an invariant mapping.
\newblock In {\em CVPR}, 2006.

\bibitem{HLJ+15}
Xufeng Han, Thomas Leung, Yangqing Jia, Rahul Sukthankar, and Alexander~C Berg.
\newblock {MatchNet}: Unifying feature and metric learning for patch-based
  matching.
\newblock In {\em CVPR}, 2015.

\bibitem{Harwood_2017_ICCV}
Ben Harwood, Vijay Kumar B~G, Gustavo Carneiro, Ian Reid, and Tom Drummond.
\newblock Smart mining for deep metric learning.
\newblock In {\em ICCV}, 2017.

\bibitem{he2019momentum}
Kaiming He, Haoqi Fan, Yuxin Wu, Saining Xie, and Ross Girshick.
\newblock Momentum contrast for unsupervised visual representation learning.
\newblock {\em arXiv preprint arXiv:1911.05722}, 2019.

\bibitem{HZRS16}
Kaiming He, Xiangyu Zhang, Shaoqing Ren, and Jian Sun.
\newblock Deep residual learning for image recognition.
\newblock In {\em CVPR}, 2016.

\bibitem{hinton2015}
Geoffrey Hinton, Oriol Vinyals, and Jeff Dean.
\newblock Distilling the knowledge in a neural network.
\newblock {\em arXiv preprint arXiv:1503.02531}, 2015.

\bibitem{hoffer2018fix}
Elad Hoffer, Itay Hubara, and Daniel Soudry.
\newblock Fix your classifier: the marginal value of training the last weight
  layer.
\newblock {\em arXiv preprint arXiv:1801.04540}, 2018.

\bibitem{howard2019searching}
Andrew Howard, Mark Sandler, Grace Chu, Liang-Chieh Chen, Bo Chen, Mingxing
  Tan, Weijun Wang, Yukun Zhu, Ruoming Pang, Vijay Vasudevan, et~al.
\newblock Searching for {MobileNetV3}.
\newblock In {\em ICCV}, 2019.

\bibitem{howard2017mobilenets}
Andrew~G Howard, Menglong Zhu, Bo Chen, Dmitry Kalenichenko, Weijun Wang,
  Tobias Weyand, Marco Andreetto, and Hartwig Adam.
\newblock {MobileNets}: Efficient convolutional neural networks for mobile
  vision applications.
\newblock {\em arXiv preprint arXiv:1704.04861}, 2017.

\bibitem{hu2019towards}
Jie Hu, Rongrong Ji, Hong Liu, Shengchuan Zhang, Cheng Deng, and Qi Tian.
\newblock Towards visual feature translation.
\newblock In {\em CVPR}, 2019.

\bibitem{huang2017}
Gao Huang, Zhuang Liu, Laurens van~der Maaten, and Kilian~Q. Weinberger.
\newblock Densely connected convolutional networks.
\newblock In {\em CVPR}, 2017.

\bibitem{huang2019gpipe}
Yanping Huang, Youlong Cheng, Ankur Bapna, Orhan Firat, Dehao Chen, Mia Chen,
  HyoukJoong Lee, Jiquan Ngiam, Quoc~V Le, Yonghui Wu, and Zhifeng Chen.
\newblock {GPipe}: Efficient training of giant neural networks using pipeline
  parallelism.
\newblock In {\em NIPS}, 2019.

\bibitem{iandola2016squeezenet}
Forrest~N Iandola, Song Han, Matthew~W Moskewicz, Khalid Ashraf, William~J
  Dally, and Kurt Keutzer.
\newblock Squeezenet: Alexnet-level accuracy with 50x fewer parameters and <0.5
  mb model size.
\newblock {\em arXiv preprint arXiv:1602.07360}, 2016.

\bibitem{Iscen_2018_CVPR}
Ahmet Iscen, Giorgos Tolias, Yannis Avrithis, and Ondřej Chum.
\newblock Mining on manifolds: Metric learning without labels.
\newblock In {\em CVPR}, 2018.

\bibitem{JJGO11}
Mihir Jain, Herve J\'egou, and Patrick Gros.
\newblock Asymmetric hamming embedding.
\newblock In {\em ACM Multimedia}, 2011.

\bibitem{JeDS10}
H. J\'egou, M. Douze, and C. Schmid.
\newblock Improving bag-of-features for large scale image search.
\newblock {\em IJCV}, 87(3), 2010.

\bibitem{JeDS11}
H. J\'egou, M. Douze, and C. Schmid.
\newblock Product quantization for nearest neighbor search.
\newblock {\em PAMI}, 33(1):117--128, 2011.

\bibitem{JeHS07}
H. J\'egou, H. Harzallah, and C. Schmid.
\newblock A contextual dissimilarity measure for accurate and efficient image
  search.
\newblock In {\em CVPR}, 2007.

\bibitem{JZ14}
Herv\'e J\'egou and Andrew Zisserman.
\newblock Triangulation embedding and democratic kernels for image search.
\newblock In {\em CVPR}, 2014.

\bibitem{KiGr10}
Jaechul Kim and Kristen Grauman.
\newblock Asymmetric region-to-image matching for comparing images with generic
  object categories.
\newblock In {\em CVPR}, 2010.

\bibitem{kim2019deep}
Sungyeon Kim, Minkyo Seo, Ivan Laptev, Minsu Cho, and Suha Kwak.
\newblock Deep metric learning beyond binary supervision.
\newblock In {\em CVPR}, 2019.

\bibitem{kiros2014unifying}
Ryan Kiros, Ruslan Salakhutdinov, and Richard~S Zemel.
\newblock Unifying visual-semantic embeddings with multimodal neural language
  models.
\newblock {\em arXiv preprint arXiv:1411.2539}, 2014.

\bibitem{LiLi15}
Shengcai Liao and Stan~Z Li.
\newblock Efficient {PSD} constrained asymmetric metric learning for person
  re-identification.
\newblock In {\em ICCV}, 2015.

\bibitem{liu2018darts}
Hanxiao Liu, Karen Simonyan, and Yiming Yang.
\newblock {DARTS}: Differentiable architecture search.
\newblock {\em arXiv preprint arXiv:1806.09055}, 2018.

\bibitem{liu2018rethinking}
Zhuang Liu, Mingjie Sun, Tinghui Zhou, Gao Huang, and Trevor Darrell.
\newblock Rethinking the value of network pruning.
\newblock In {\em ICLR}, 2018.

\bibitem{Lowe04}
D.G. Lowe.
\newblock Distinctive image features from scale-invariant keypoints.
\newblock {\em IJCV}, 60(2):91--110, 2004.

\bibitem{misra2019self}
Ishan Misra and Laurens van~der Maaten.
\newblock Self-supervised learning of pretext-invariant representations.
\newblock {\em arXiv preprint arXiv:1912.01991}, 2019.

\bibitem{NSS+13}
Behnam Neyshabur, Nati Srebro, Ruslan~R Salakhutdinov, Yury Makarychev, and
  Payman Yadollahpour.
\newblock The power of asymmetry in binary hashing.
\newblock In {\em NIPS}, 2013.

\bibitem{OXJS16}
Hyun Oh~Song, Yu Xiang, Stefanie Jegelka, and Silvio Savarese.
\newblock Deep metric learning via lifted structured feature embedding.
\newblock In {\em CVPR}, 2016.

\bibitem{park2019relational}
Wonpyo Park, Dongju Kim, Yan Lu, and Minsu Cho.
\newblock Relational knowledge distillation.
\newblock In {\em CVPR}, 2019.

\bibitem{pham2018efficient}
Hieu Pham, Melody~Y Guan, Barret Zoph, Quoc~V Le, and Jeff Dean.
\newblock Efficient neural architecture search via parameter sharing.
\newblock {\em arXiv preprint arXiv:1802.03268}, 2018.

\bibitem{PCISZ07}
J. Philbin, O. Chum, M. Isard, J. Sivic, and A. Zisserman.
\newblock Object retrieval with large vocabularies and fast spatial matching.
\newblock In {\em CVPR}, 2007.

\bibitem{PCS+08}
James Philbin, Ondrej Chum, Josef Sivic, Michael Isard, and Andrew Zisserman.
\newblock Lost in quantization: Improving particular object retrieval in large
  scale image databases.
\newblock In {\em CVPR}, 2008.

\bibitem{RIT+18}
Filip Radenovi{\'c}, Ahmet Iscen, Giorgos Tolias, Yannis Avrithis, and
  Ond{\v{r}}ej Chum.
\newblock Revisiting oxford and paris: Large-scale image retrieval
  benchmarking.
\newblock In {\em CVPR}, 2018.

\bibitem{RTC16}
Filip Radenovi{\'c}, Giorgos Tolias, and Ond{\v{r}}ej Chum.
\newblock {CNN} image retrieval learns from bow: Unsupervised fine-tuning with
  hard examples.
\newblock {\em ECCV}, 2016.

\bibitem{radenovic2018fine}
Filip Radenovi{\'c}, Giorgos Tolias, and Ond{\v{r}}ej Chum.
\newblock Fine-tuning {CNN} image retrieval with no human annotation.
\newblock {\em PAMI}, 41(7):1655--1668, 2018.

\bibitem{RDG+18}
Ilija Radosavovic, Piotr Dollar, Ross Girshick, Georgia Gkioxari, and Kaiming
  He.
\newblock Data distillation: Towards omni-supervised learning.
\newblock In {\em CVPR}, 2018.

\bibitem{RSMC14}
Ali~Sharif Razavian, Josephine Sullivan, Atsuto Maki, and Stefan Carlsson.
\newblock Visual instance retrieval with deep convolutional networks.
\newblock {\em arXiv preprint arXiv:1412.6574}, 2014.

\bibitem{russakovsky2015imagenet}
Olga Russakovsky, Jia Deng, Hao Su, Jonathan Krause, Sanjeev Satheesh, Sean Ma,
  Zhiheng Huang, Andrej Karpathy, Aditya Khosla, Michael Bernstein, et~al.
\newblock {ImageNet} large scale visual recognition challenge.
\newblock {\em IJCV}, 115(3):211--252, 2015.

\bibitem{sandler2018mobilenetv2}
Mark Sandler, Andrew Howard, Menglong Zhu, Andrey Zhmoginov, and Liang-Chieh
  Chen.
\newblock Mobilenetv2: Inverted residuals and linear bottlenecks.
\newblock In {\em CVPR}, 2018.

\bibitem{scholkopf1998nonlinear}
Bernhard Sch{\"o}lkopf, Alexander Smola, and Klaus-Robert M{\"u}ller.
\newblock Nonlinear component analysis as a kernel eigenvalue problem.
\newblock {\em Neural computation}, 10(5):1299--1319, 1998.

\bibitem{shen2020towards}
Yantao Shen, Yuanjun Xiong, Wei Xia, and Stefano Soatto.
\newblock Towards backward-compatible representation learning.
\newblock {\em arXiv preprint arXiv:2003.11942}, 2020.

\bibitem{simonyan2014very}
Karen Simonyan and Andrew Zisserman.
\newblock Very deep convolutional networks for large-scale image recognition.
\newblock {\em arXiv preprint arXiv:1409.1556}, 2014.

\bibitem{snell2017}
Jake Snell, Kevin Swersky, and Richard Zemel.
\newblock Prototypical networks for few-shot learning.
\newblock In {\em NIPS}, 2017.

\bibitem{tan2019mnasnet}
Mingxing Tan, Bo Chen, Ruoming Pang, Vijay Vasudevan, Mark Sandler, Andrew
  Howard, and Quoc~V Le.
\newblock {MnasNet}: Platform-aware neural architecture search for mobile.
\newblock In {\em CVPR}, 2019.

\bibitem{tan2019efficientnet}
Mingxing Tan and Quoc~V Le.
\newblock {EfficientNet}: Rethinking model scaling for convolutional neural
  networks.
\newblock {\em arXiv preprint arXiv:1905.11946}, 2019.

\bibitem{Tenenbaum97}
J. Tenenbaum.
\newblock Mapping a manifold of perceptual observations.
\newblock In {\em NIPS}. 1997.

\bibitem{ToAJ13}
Giorgos Tolias, Yannis Avrithis, and Herv\'e J\'egou.
\newblock To aggregate or not to aggregate: Selective match kernels for image
  search.
\newblock In {\em ICCV}, 2013.

\bibitem{tolias2017asymmetric}
Giorgos Tolias and Ondrej Chum.
\newblock Asymmetric feature maps with application to sketch based retrieval.
\newblock In {\em CVPR}, 2017.

\bibitem{TSJ15}
Giorgos Tolias, Ronan Sicre, and Herv{\'e} J{\'e}gou.
\newblock Particular object retrieval with integral max-pooling of cnn
  activations.
\newblock {\em ICLR}, 2016.

\bibitem{vinyals2016}
Oriol Vinyals, Charles Blundell, Timothy Lillicrap, Koray Kavu\-kcuoglu, and
  Daan Wierstra.
\newblock Matching networks for one shot learning.
\newblock In {\em NIPS}, 2016.

\bibitem{wang2018cosface}
Hao Wang, Yitong Wang, Zheng Zhou, Xing Ji, Dihong Gong, Jingchao Zhou, Zhifeng
  Li, and Wei Liu.
\newblock {CosFace}: Large margin cosine loss for deep face recognition.
\newblock In {\em CVPR}, 2018.

\bibitem{WSL+14}
Jiang Wang, Yang Song, Thomas Leung, Chuck Rosenberg, Jingbin Wang, James
  Philbin, Bo Chen, and Ying Wu.
\newblock Learning fine-grained image similarity with deep ranking.
\newblock In {\em CVPR}, 2014.

\bibitem{WHH+19}
Xun Wang, Xintong Han, Weilin Huang, Dengke Dong, and Matthew~R Scott.
\newblock Multi-similarity loss with general pair weighting for deep metric
  learning.
\newblock In {\em CVPR}, 2019.

\bibitem{wang2019cross}
Xun Wang, Haozhi Zhang, Weilin Huang, and Matthew~R Scott.
\newblock Cross-batch memory for embedding learning.
\newblock {\em arXiv preprint arXiv:1912.06798}, 2019.

\bibitem{wang2020progressive}
Xiaodong Wang, Zhedong Zheng, Yang He, Fei Yan, Zhiqiang Zeng, and Yi Yang.
\newblock Progressive local filter pruning for image retrieval acceleration.
\newblock {\em arXiv preprint arXiv:2001.08878}, 2020.

\bibitem{Wu_2017_ICCV}
Chao-Yuan Wu, R. Manmatha, Alexander~J. Smola, and Philipp Krahenbuhl.
\newblock Sampling matters in deep embedding learning.
\newblock In {\em ICCV}, 2017.

\bibitem{wu2018improving}
Zhirong Wu, Alexei~A Efros, and Stella~X Yu.
\newblock Improving generalization via scalable neighborhood component
  analysis.
\newblock In {\em ECCV}, 2018.

\bibitem{xia2008listwise}
Fen Xia, Tie-Yan Liu, Jue Wang, Wensheng Zhang, and Hang Li.
\newblock Listwise approach to learning to rank: theory and algorithm.
\newblock In {\em ICML}, 2008.

\bibitem{XJRN03}
Eric~P Xing, Michael~I Jordan, Stuart~J Russell, and Andrew~Y Ng.
\newblock Distance metric learning with application to clustering with
  side-information.
\newblock In {\em NIPS}, 2003.

\bibitem{XYDZ19}
Xinyi Xu, Yanhua Yang, Cheng Deng, and Feng Zheng.
\newblock Deep asymmetric metric learning via rich relationship mining.
\newblock In {\em CVPR}, 2019.

\bibitem{Ye_2019_CVPR}
Mang Ye, Xu Zhang, Pong~C. Yuen, and Shih-Fu Chang.
\newblock Unsupervised embedding learning via invariant and spreading instance
  feature.
\newblock In {\em CVPR}, June 2019.

\bibitem{DBLP:conf/iccv/YuWZ17}
Hong{-}Xing Yu, Ancong Wu, and Wei{-}Shi Zheng.
\newblock Cross-view asymmetric metric learning for unsupervised person
  re-identification.
\newblock In {\em ICCV}, 2017.

\bibitem{YYL+19}
Lu Yu, Vacit~Oguz Yazici, Xialei Liu, Joost Van~De Weijer, Yongmei Cheng, and
  Arnau Ramisa.
\newblock Learning metrics from teachers: Compact networks for image embedding.
\newblock In {\em CVPR}, 2019.

\bibitem{zagoruyko2016wide}
Sergey Zagoruyko and Nikos Komodakis.
\newblock Wide residual networks.
\newblock {\em arXiv preprint arXiv:1605.07146}, 2016.

\bibitem{zhang2018shufflenet}
Xiangyu Zhang, Xinyu Zhou, Mengxiao Lin, and Jian Sun.
\newblock {ShuffleNet}: An extremely efficient convolutional neural network for
  mobile devices.
\newblock In {\em CVPR}, 2018.

\bibitem{ZhJI13}
Cai-Zhi Zhu, Herve J\'egou, and Shin'ichi Satoh.
\newblock Query-adaptive asymmetrical dissimilarities for visual object
  retrieval.
\newblock In {\em ICCV}, 2013.

\end{thebibliography}
